\documentclass[]{iopjournal}

\usepackage{booktabs}
\usepackage{multicol}
\usepackage{float}
\usepackage{graphicx}
\usepackage{amsmath}
\usepackage{amsfonts}
\usepackage{amssymb}
\usepackage{subcaption}
\usepackage{adjustbox}
\usepackage{algorithm}
\usepackage{algpseudocode}
\usepackage{appendix}
\usepackage{tikz}
\usepackage{xcolor}
\usetikzlibrary{patterns}
\usepackage[T1]{fontenc}
\usepackage[protrusion=true, expansion=true]{microtype}
\usepackage{xurl}
\usepackage[english]{babel}

\usepackage[colorlinks=true,allcolors=blue]{hyperref}

\usepackage[style=numeric, natbib=true, maxbibnames=10]{biblatex}
\newcommand{\citetnum}[1]{\citeauthor{#1}~\cite{#1}}
\AtEveryBibitem{%
  \clearfield{issn}   %
  \clearfield{isbn}   %
  \clearfield{month}
  \clearfield{language}
}

\definecolor{bayes}{HTML}{009E73}
\definecolor{density}{HTML}{E69F00}
\definecolor{distance}{HTML}{0072B2}
\definecolor{baseline}{HTML}{8C8C8C}

\DeclareRobustCommand{\legendbox}[2]{%
  \begingroup
  \setlength{\fboxsep}{0.5pt}%
  \colorbox{#1}{\textcolor{white}{\strut\kern1.5pt #2\kern1.5pt}}%
  \endgroup
}

\DeclareRobustCommand{\tikzboxthree}{%
  \tikz[baseline=0em]{\draw (0,0) rectangle (1em,0.5em);}%
}
\DeclareRobustCommand{\tikzboxtwo}{%
  \tikz[baseline=0em]{\draw[fill=white,pattern=north east lines,pattern color=black] (0,0) rectangle (1em,0.5em);}%
}

\usepackage[symbol]{footmisc}

\begin{document}
\onecolumn

\title{The Challenge of Out-Of-Distribution Detection in Motor Imagery BCIs}

\author{Merlijn Quincent Mulder $^1$\orcid{0009-0002-0880-1059}, Matias Valdenegro-Toro$^{1}$\orcid{0000-0001-5793-9498}, Andreea Ioana Sburlea$^{1}\footnote{indicates shared last author.}
$\orcid{0000-0001-6766-3464} and Ivo Pascal de Jong $^{1*}$\orcid{0000-0002-1497-8013}}

\affil{$^1$Department of Artificial Intelligence, University of Groningen \\9747 AG Groningen, The Netherlands}

\email{ivo.de.jong@rug.nl}


\publication{Merlijn Quincent Mulder et al. 2026,
\textit{J. Neural Eng.} \textbf{23}(5), 056046.
\href{https://doi.org/10.1088/1741-2552/aeac87}
{10.1088/1741-2552/aeac87}}

\begin{abstract}

\subsection*{Objective}
 Machine Learning classifiers used in Brain-Computer Interfaces make classifications based on the distribution of examples on which they were trained. When exposed to EEG from unfamiliar classes, they can only make blind guesses. Instead of allowing such guesses, these Out-of-Distribution (OOD) samples should be detected and rejected to improve the robustness of BCIs against unfamiliar cognitive states.

\subsection*{Approach}
We study OOD detection in Motor Imagery BCIs by training a model on some classes and observing whether an unfamiliar movement class can be detected based on increased uncertainty. We tested seven different OOD detection methods and one more method that has been claimed to boost the quality of OOD detection. 

\subsection*{Main results}
 For many BCI users, the uncertainty for the familiar in-distribution classes can still be higher than for the out-of-distribution classes, due to the high intrinsic variability inherent in EEG signals. 
 As a result, many OOD detection methods that have shown good performance in other machine learning domains prove to be ineffective at identifying unfamiliar motor imagery patterns in BCIs. However, we also found that OOD detection performance is correlated with on-task performance, and that Deep Ensemble models and MC-Dropout models were able to achieve on-task AUROC $>$ 0.9, and OOD detection ability up to 0.7. This shows that for models and subjects where task performance is high, rejecting unfamiliar cognitive states becomes feasible.

\subsection*{Significance}
Our research demonstrates a Leave-One-Class-Out OOD detection setup as an experimental paradigm for studying whether models are robust against unfamiliar motor imagery actions that were not seen in the training data. This is the first benchmark in Motor Imagery BCI that evaluates this type of OOD data. Our results demonstrate how to improve the overall safety and reliability of BCIs by preventing erroneous actions.

\end{abstract}

\twocolumn

\section{Introduction}\label{sec:introduction}

Machine Learning models used in Brain-Computer Interfaces (BCIs) make mistakes when evaluated on data that is different from the training data.
\citet{EEGShifts} have shown that under distribution shifts in electroencephalography (EEG) data, model performance degrades rapidly. When a model is used on data that is different from the training data, these examples are considered \textit{out-of-distribution} (OOD). A good model that generalises well is robust against small distribution shifts, but some shifts are so large that it is impossible to make good classifications, and the only solution is to detect that these samples are OOD and abstain from classifying them.

\begin{figure}[t]
    \centering
    \includegraphics[width=1\linewidth]{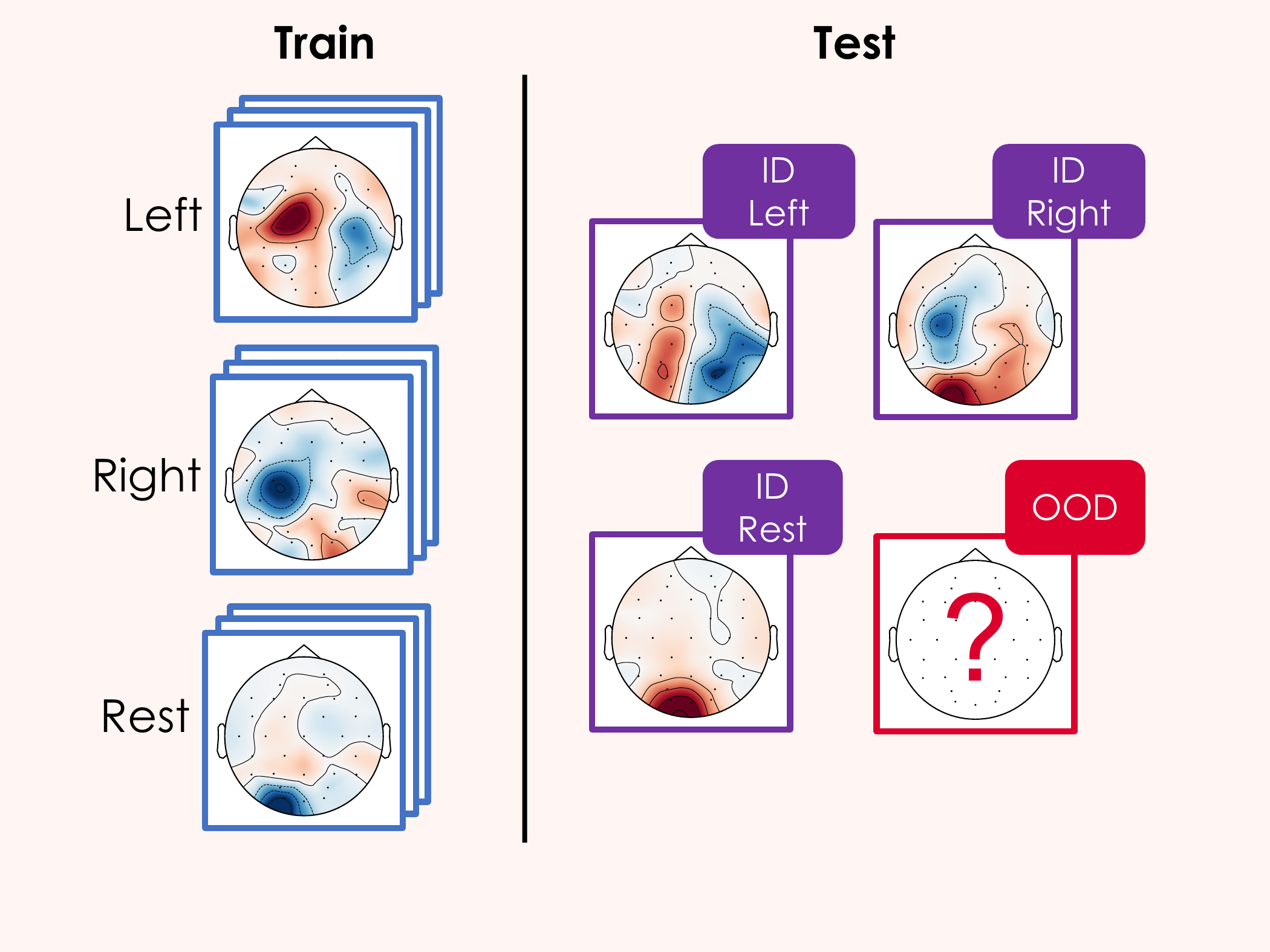}
    \caption{Concept of out-of-distribution detection. A model is trained on several known classes. It must be able to recognise when unknown data do not belong to any known class, while still classifying the known classes correctly.}
    \label{fig:intro}
\end{figure}

Models in Motor Imagery BCIs are typically trained and evaluated with EEG data from two (e.g. left-hand, right-hand) to five (e.g. left, right, foot, tongue, rest) classes and are deployed with the assumption that all signals later must match one of these classes. In reality, it is very likely that users will perform actions that do not match any of these classes at some point during usage \cite{BNN_OOD}. In this case, the meaning of the signal is different from that of the samples that the model previously observed. This is considered a \textit{semantic shift}, where a test sample belongs to a class on which the model was never trained, as shown in Figure \ref{fig:intro}.

In this case, it is impossible to get a correct classification, as none of the known classes is correct. The solution to these samples is to use OOD-detection methods to identify them as unfamiliar and abstain from making a classification.

\begin{figure*}[!t]
    \centering
    \includegraphics[width=\linewidth]{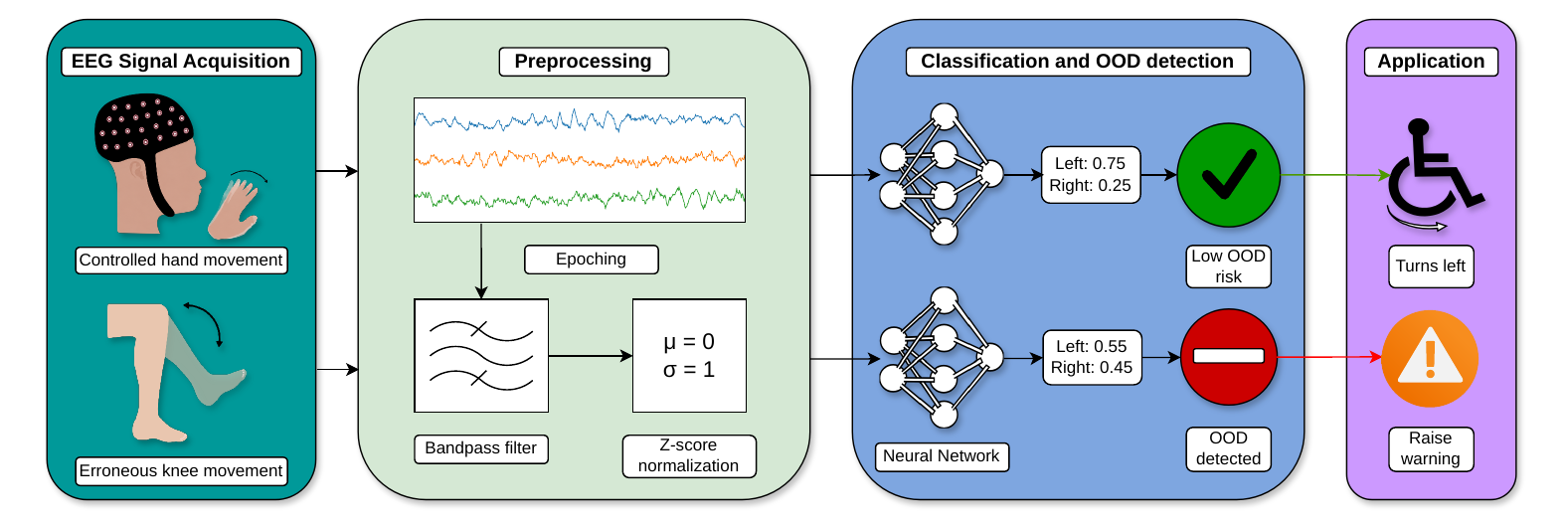}
    \caption{A BCI system with OOD detection. When the user executes an unfamiliar movement, the Machine Learning model identifies this as out-of-distribution, and the system refuses to execute an incorrect command. This improves safety when users show off-task behaviour. 
    }
    \label{fig:BCIUQloop}
\end{figure*}

OOD detection works by using Uncertainty Quantification (UQ) methods that aim to give robust estimates of whether a classification is likely to be correct \cite{de2023uncertainty, yang_generalized_2024}. By indicating high uncertainty for OOD samples and low uncertainty estimates for in-distribution (ID) samples from the known classes, the model can indicate whether a sample is likely to be OOD or not. When a sample is OOD, the BCI pipeline should classify it as ``unfamiliar'' instead of incorrectly guessing one of the known classes. An example of a BCI pipeline that supports OOD detection is illustrated in Figure \ref{fig:BCIUQloop}.

OOD detection as a topic has received only limited attention in BCI research \cite{BNN_OOD, jong_how_2024}, but in the broader Machine Learning literature, this topic is already extensively researched \cite{OODTrend}. An extensive review of dedicated OOD methods and perspectives is described in\citet{yang_generalized_2024}. \citet{mucsanyi_benchmarking_2024} compares many of these methods on their capabilities and shows that these different methods yield varying performance across tasks. Hence, there is a need for task-specific approaches when using uncertainty estimation methods for OOD detection. Notably, most methods have been developed for image classification problems and cannot be applied straightaway to other domains with the same promise of performance. In particular, EEG is known to have large amounts of noise, which can generally lead to higher uncertainty estimates for ID data, which could make OOD detection more difficult \cite{jong_how_2024}. 

While there has already been some research on estimating uncertainty in BCIs \citep{UNCER, BayesianNetworksBCI, Milanes2021, Kim2026, suurmeijer2025uncertainty, diffMCvsNormal}, evaluating uncertainty to detect unfamiliar Motor Imagery movements as Out-of-Distribution has been minimally explored, yielding performances marginally above random guessing \citep{jong_how_2024}.

In this study, we establish the state of the art of OOD detection for BCIs, exploring a wide range of OOD detection methods for Convolutional Neural Networks (CNN) used in BCIs to identify which methods have the greatest impact on improving BCI robustness. We also investigated which factors have an impact on OOD detection performance, looking at in-distribution classification performance, specific classes that may be easier or harder to detect, and the activation strength of the penultimate layer in the neural network. Through this, we establish the current state-of-the-art, explain why OOD detection is particularly difficult in BCIs, and show how OOD detection may improve with on-task performance. 

To this end, we benchmark seven OOD detection methods that span Bayesian Neural Network, distance-based, and density-based approaches as defined in Section \ref{sec:ood}. We will use three different datasets (from \cite{BCNI2014, schirrmeister_deep_2017, stieger2021}), available through Mother of all BCI Benchmarks (MOABB) \cite{MOABB}. 
As our CNN, we use EEGNeX, an optimised version of EEGNet that proved a reliable EEG signal decoding CNN-based architecture \cite{chen_toward_2024}.

\subsection{Contributions}

By studying OOD detection in Motor Imagery BCIs, we make the following contributions:

\begin{itemize}
    \item We demonstrate leave-one-class-out OOD detection as a realistic and challenging setup for BCIs, showing that it is possible to detect these samples with performance better than random guessing, contrary to previous findings \cite{jong_how_2024}.
    
    \item We benchmark known OOD methods from Computer Vision and find that they do not necessarily transfer well to BCIs. However, we established Monte Carlo (MC) Dropout and Deep Ensembles as the most effective approaches to improve BCI robustness in the first leave-one-class-out OOD detection benchmark in MI-BCI.

    \item We show that the performance of the OOD detector correlates with the on-task classification performance. This means that OOD detection will get better as classification improves.

\end{itemize}

We demonstrate that for some datasets and specific OOD detection methods, OOD detection performs quite well (AUROC up to 0.7), in particular for users where in-distribution task performance is also high. We conclude that for users with reliable BCI use, these methods can be used to detect when the user performs off-task behaviour and abstain from executing an unintended command. Through this, we show that OOD detection has the potential to improve the robustness and safety of BCIs. 

\subsection{Leave-one-class-out OOD Detection (LOCO-OOD)}\label{sec:locoood}

We propose Leave-One-Class-Out OOD Detection (LOCO-OOD) as an experimental paradigm to evaluate the performance of OOD detection models for BCIs that can be applied to existing datasets, but is ecologically valid. 

LOCO-OOD trains models on a subset of classes available in the training dataset and evaluates them on all classes. That means that there is one unseen class in the test dataset. A good OOD detection model should report higher uncertainty in the unseen class while still giving good predictions on the familiar classes. By applying a varying threshold on the estimated uncertainty, the unseen OOD class should be separated. The degree of separation of the OOD class can be measured with the Area Under the Receiver-Operator Characteristic curve (AUROC). The procedure for LOCO-OOD experiments is specified in Algorithm \ref{alg:loco}.

LOCO-OOD represents OOD samples that arise due to the off-task behaviour of the user. The model is trained on the data of some classes, but when the user does something unfamiliar, the sample should be rejected. A model that performs well LOCO-OOD detection is robust against these errors. This is equivalent to a user performing a calibration session with several classes, but then during testing exhibiting behaviour that does not fit any of these classes.

Note that this setup only tests for users exhibiting off-task behaviour. It does not deal with drifts, artifacts or other causes of uncertainty. These are common causes of uncertainty that should be addressed with artifact removal \citep{susan2023systematic} or unsupervised recalibration \citep{wilson2025long}, which fall outside the scope of the current work. 

\begin{algorithm}[t]
\caption{LOCO-OOD experiment }\label{alg:loco}
\begin{algorithmic}[1]
\State $hyperparameters \gets tune(subjects[0])$

\For{$subject$ in subjects}

    \For{OOD-class in classes}
        \State $train \gets \textsc{get\_train}(subject)$
        \Statex \hspace{\algorithmicindent} \Comment{excludes OOD-class}

        \State $test \gets \textsc{get\_test(subject)}$
        \Statex \hspace{\algorithmicindent} \Comment{includes OOD-class}
        \State $model \gets \textsc{EEGNeX()}$
        \State \textsc{train($model, train$)}

        \State $\textsc{AUROC} \gets \textsc{OOD}(model, test)$
    \EndFor
\EndFor

\end{algorithmic}
\end{algorithm}

\section{Background}

In this section, we provide relevant background information. We introduce fundamental concepts of uncertainty, briefly recap the essentials of Motor Imagery BCIs and discuss related works. 

Following this, we extensively document existing OOD detection methods and provide a conceptual categorisation. This serves as a review of the benchmarked OOD detection methods, aiming to provide the reader with an intuition of how these methods work and how they differ.

\subsection{Preliminaries}

As defined by \citetnum{yang_generalized_2024}, a trustworthy recognition system is one that not only makes accurate predictions on known in-distribution (ID) data but can also reliably detect and reject samples outside of this distribution. Within this article, we consider a specific case of OOD detection where we define OOD samples as unfamiliar classes represented by unfamiliar Motor Imagery actions. OOD detection can be more broadly approached to consider unfamiliar examples of a known class, for example, due to unseen artifacts and slow drifts, but under those conditions, we argue that the model should generalise, rather than reject the samples. Prior work has explored this generalisation to OOD conditions through EEG-foundation models \cite{zheng2025beatrix}.

\paragraph{Uncertainty}
Current OOD detection methods rely on uncertainty estimators to distinguish between ID and OOD samples. This follows from the assumption that a model should produce a prediction on a data point from an unknown distribution with low confidence. However, this is not always trivial \cite{nguyen_deep_2015}. Typically, studies separate uncertainty into two types \cite{wang_aleatoric_2025}:

\begin{enumerate}
    \item \textit{Aleatoric uncertainty} is inherent to noise and ambiguity in the data, leading to low-confidence predictions. Aleatoric uncertainty is a consequence of the task not being perfectly `solvable', leading to lower classification accuracy \cite{li_prototype-based_2024}. If aleatoric uncertainty is high in ID samples, it may interfere with OOD detection.
    \item \textit{Epistemic uncertainty} can be described as the lack of knowledge in the model. In the context of OOD samples belonging to different classes, this is a quantity that is high for a previously unseen input coming from a different distribution than the class distributions the model learned.
\end{enumerate}

Correctly and truly disentangling the two types of uncertainty yields more reliable OOD detection under the previously mentioned definition of OOD. Currently, we do not have a good method for disentanglement \cite{jong_how_2024}. 
A robust model will account for both uncertainties. We will focus on detecting OOD samples using proven methods that do not always explicitly distinguish between the types of uncertainty.

\paragraph{MI-BCI paradigm} In motor rehabilitation, BCIs are used to reintegrate the sensory-motor loop by directly accessing brain activity  \cite{alonso-valerdi_motor_2015}. These motor interfaces translate neural signals into device commands, often relying on an endogenous task. Such tasks often require skills and experience to produce well-distinguishable brain patterns for accurate classification. The patterns are often recorded through EEG. Using differential amplification, waveforms are generated by comparing electrodes on the scalp, measuring voltage over time \cite{louis_orderly_2016}. 

Typically, muscle activity is associated with Mu (8-12 Hz) and Beta (13-25 Hz) waves \cite{louis_orderly_2016}. However, because EEG signals are generally weak, they are prone to artifacts that introduce noise before reaching the scalp. These noisy signals are later transformed into features that encode the user's task, from which control commands can be derived.

\begin{figure*}[t]
    \centering
    \begin{subfigure}[t]{0.3\textwidth}
        \centering
        \includegraphics[width=\linewidth]{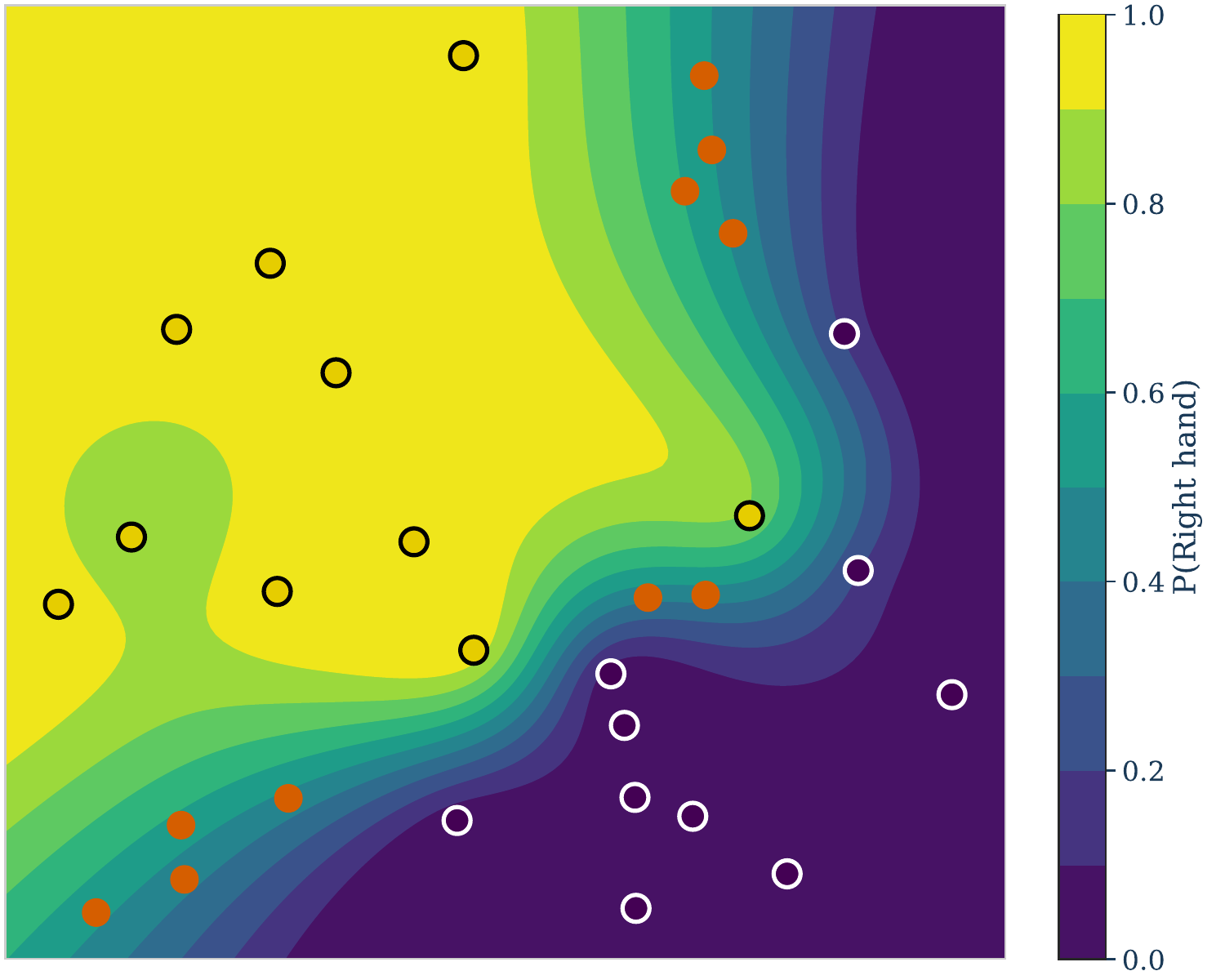}
        \caption{Bayesian Neural Network methods. OOD samples are expected to be in a region with a low or equal probability for all classes.}
        \label{fig:img1}
    \end{subfigure}
    \hfill
    \begin{subfigure}[t]{0.3\textwidth}
        \centering
        \includegraphics[width=\linewidth]{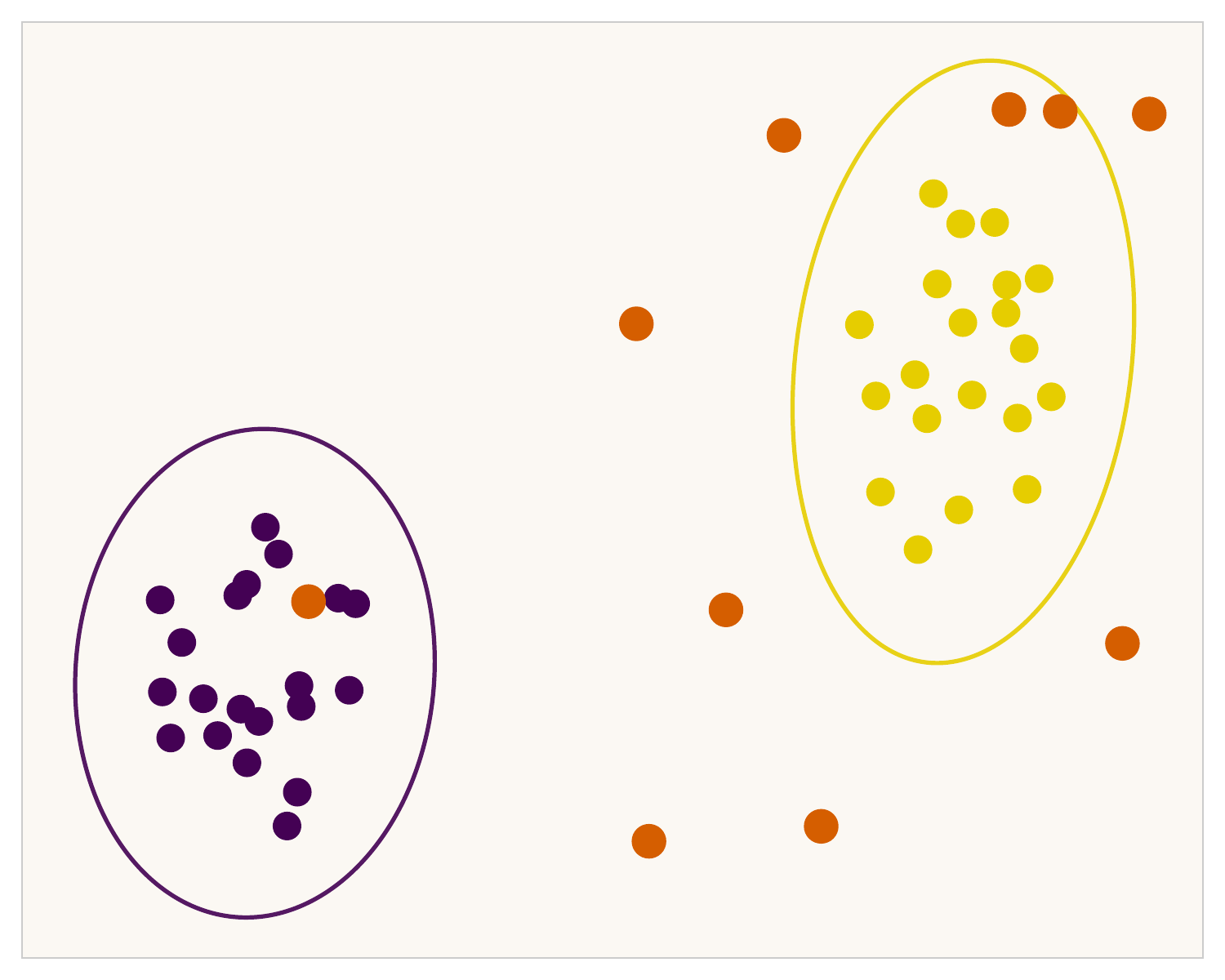}
        \caption{Density-based methods. A distribution of ID samples is learned, and samples that are far away from the distribution are expected to be OOD.}
        \label{fig:img2}
    \end{subfigure}
    \hfill
    \begin{subfigure}[t]{0.3\textwidth}
        \centering
        \includegraphics[width=\linewidth]{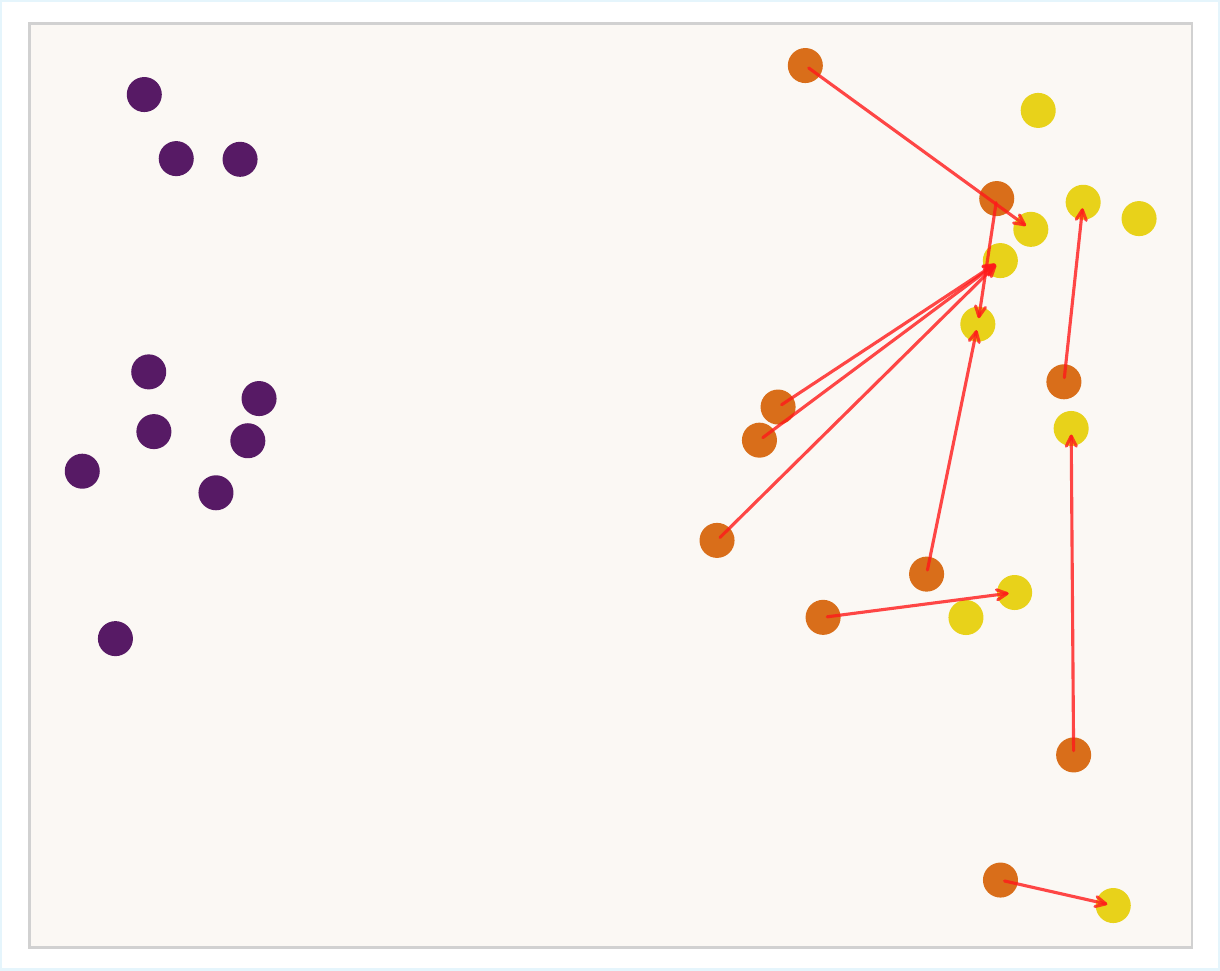}
        \caption{Distance-based methods. Neighbours from the same ID class are close together, whereas samples that are far away from ID samples are considered OOD.}
        \label{fig:img3}
    \end{subfigure}
    \caption{A visual explanation of three different OOD detection approaches. The ID data points belong to user nine of the BNCI2014 dataset. In this illustration 
    \colorbox[HTML]{D55E00}{\textcolor{white}{feet}} is considered OOD on a model trained on the \colorbox[HTML]{E6CD00}{\textcolor{black}{right hand}} and \colorbox[HTML]{440154}{\textcolor{white}{left hand}} classes.
    \label{fig:OOD_detection_approaches}}
\end{figure*}

\paragraph{Related works}
Prior work has proposed to estimate and reduce uncertainty in neural decoding of ID EEG signals through adaptive mixing of augmentations on the input sequence, \cite{UNCER}, but does not extend this to OOD input.
Others propose a threshold-based approach using uncertainty to prevent making predictions on noisy and uncertain ID signals at all in order to increase the reliability of neural decoding and classification approaches \cite{BayesianNetworksBCI, Milanes2021, Kim2026}. 

Besides abstaining on uncertain predictions on ID data, a reliable model should also extend this to OOD data.
There has already been some interest in the task of OOD detection in EEG. \citet{wong_estimating_2023} studied uncertainty and OOD detection in epilepsy and proposed a novel anomaly detection method using the description of deep support vector data and the uncertainty of the model, demonstrating an increase in the robustness of a model trained for seizure detection. However, they used OOD detection to identify patients who were different from the training data, for whom the model would not work well. Detecting samples within-subject that are OOD requires per-sample OOD detection. \citet{wagh_evaluating_2022} extensively studied the estimation of uncertainty under realistic EEG distribution shifts based on simulated instrumentation-related variabilities, but not brain-related variabilities, nor inference under unfamiliar classes. 

\citet{BNN_OOD} focused on Bayesian approaches for OOD detection in BCIs, though with a different way to represent OOD samples that does not have an immediate ecological equivalent. They conclude that methods developed in other domains fail to transfer to BCI tasks. Previous literature has not established a state of the art for OOD detection in BCIs, nor does it identify factors that explain its performance.

Recently, \citet{liu2026} proposed an OOD detection framework for asynchronous MI-BCI operating on continuous EEG streams, introducing a temporal density OOD score. They formulate online control as a hierarchical decision process in which a first-stage model performs rest/task gating, followed by a second-stage model for ID classification and OOD rejection. Their experimental evaluation treats left- and right-hand motor imagery as the in-distribution classes, while held-out motor imagery classes are considered OOD.

\subsection{OOD Detection Approaches}\label{sec:ood}

We consider three distinct categories of OOD detection approaches, which are visualised in Figure \ref{fig:OOD_detection_approaches}. These categories offer an overview of the high-level concepts underlying the OOD detection methods that are tested.

\subsubsection{Bayesian Neural Network}
Bayesian Neural Networks represent the parameters of the model as a probability distribution estimated using Bayesian statistics. This differs from classical neural networks that use point-wise parameters. Bayesian Neural Networks represent uncertainty through disagreement among models, enabling them to estimate epistemic uncertainty. We implement Bayesian Neural Networks as MC Dropout \citep{gal_dropout_2016} and Deep Ensemble \citep{deep_ensembles}, as they have generally been shown to perform well \cite{mucsanyi_benchmarking_2024} and do not require altering the model structure and training process of the feature extractor.

\paragraph{Monte Carlo (MC) Dropout}
 \citet{gal_dropout_2016} proposed to randomly remove neurons in dropout layers during inference according to a predetermined dropout probability. 
 Through multiple forward passes with the same samples, a set of predictions is obtained corresponding to different configurations of the model.
 We measure the uncertainty using the entropy of the mean probabilities.
This method can be applied to any trained model that includes a dropout layer.

{%
    \setlength{\emergencystretch}{3em}%
    \paragraph{Deep Ensemble (DE)} 
    \citet{deep_ensembles} proposed aggregating predictions from multiple independently trained models that share the same architecture but differ in weight initialisation. Because each model converges to a different local optimum, the ensemble can be interpreted as approximating samples from the posterior distribution over model parameters. In addition to yielding good OOD detection abilities, this can also improve in-distribution classification performance \citep{diffMCvsNormal}. 

}

\subsubsection{Density Based}
Density-based methods learn the ID distribution, quantifying the likelihood that an input is ID. Samples that have a very low likelihood of belonging to the ID distribution are considered OOD.

\paragraph{Deep Deterministic Uncertainty (DDU)} 
\citet{mukhoti_deep_2023} proposed using Gaussian Discriminant Analysis to estimate epistemic uncertainty.
They fit a Gaussian Mixture Model (GMM) with class-specific covariance matrices to the embeddings from the penultimate layer.
A sample is considered OOD when its estimated probability density under the fitted GMM is lower than that of ID samples.
The authors argue that DDU can better separate uncertainties than other methods (DUQ, \cite{amersfoort_uncertainty_2020}), especially on ambiguous datasets. As EEG data is often noisy and ambiguous, this can be an advantage.

\paragraph{Energy Score}
\citet{liu_energy-based_2021} proposed the energy score as an alternative to the softmax class probabilities. Theoretically, it aligns with the probability density of the inputs, being less susceptible to overconfidence, and thereby improving distribution separation. Each input is mapped to a scalar using the energy function, which is lower for observed data than for unobserved data. 
A sample is classified as OOD if its energy score exceeds a threshold, where the threshold is determined based on the energy scores of ID samples.
Although not explicitly probabilistic, the energy score can be considered an implicit density method, as it can be extended to a probabilistic model using the Gibbs distribution \cite{energy_tutorial} and relies on the assumption that higher density regions are more likely.

\subsubsection{Distance Based}

Distance-based UQ methods determine whether an example is OOD by measuring the distance between a given input and the ID training data in the latent space. These methods assume that samples relatively far from the ID data are more likely to be OOD.

\paragraph{Deterministic Uncertainty Quantification (DUQ)} 

Proposed by \citet{amersfoort_uncertainty_2020}, DUQ computes the exponentiated distance between a feature embedding and centroids corresponding to each class. Using a similarity score, they quantified the model's uncertainty, where the similarity is higher when a feature is closer to the class centroid in the latent space. When the similarity to all centroids is low and the distance is relatively large, a sample can be considered OOD. 
DUQ is composed of a (pre-trained) deep model used as a feature extractor and a radial basis function (RBF) network. 
The kernel allows for feature insensitivity per class, which minimises the potential for feature collapse and leads to stable training.

\paragraph{Deep \textit{k}-Nearest Neighbour (d-KNN)}
\citet{sun_out--distribution_2022} propose a method based on the \textit{k}-nearest neighbour distance from the feature embeddings.
For all training samples, the L2-normalised feature embeddings are extracted and stored. When a new input sample is presented, its feature embedding will be compared to all stored embeddings using Euclidean distance. From the sorted distances, the \textit{k}-th nearest neighbour is found. 
If the distance for a sample is relatively large compared to most ID samples, it can be considered OOD. 
d-KNN's advantage over DUQ is that d-KNN is less computationally expensive, as it does not require training an additional model head with its own hyperparameters on top of the feature extractor.

\subsubsection{OOD detection Optimisation}
In addition to evaluating OOD detection methods, this study investigates Rectified Activations (ReAct) as a general OOD optimisation technique to enhance OOD detection performance. This technique is designed to be used in conjunction with other approaches and can be applied across all OOD detection methods.

\paragraph{Rectified Activations (ReAct)}
OOD inputs can cause activation patterns in hidden layers of the Neural Network that deviate significantly from those produced by ID data, as shown in image classification tasks \cite{sun_react_2021}. This results in larger variations across activations biased towards having high values, leading to model overconfidence. ReAct truncates activations above a limit $c$ to limit the effect of noise. This variable is set based on the p-th percentile of activations estimated on ID data. Combining this technique with d-KNN or energy-based OOD has previously shown improved performance \cite{sun_react_2021}.

\begin{figure*}[!t]
    \centering
    \includegraphics[width=\linewidth]{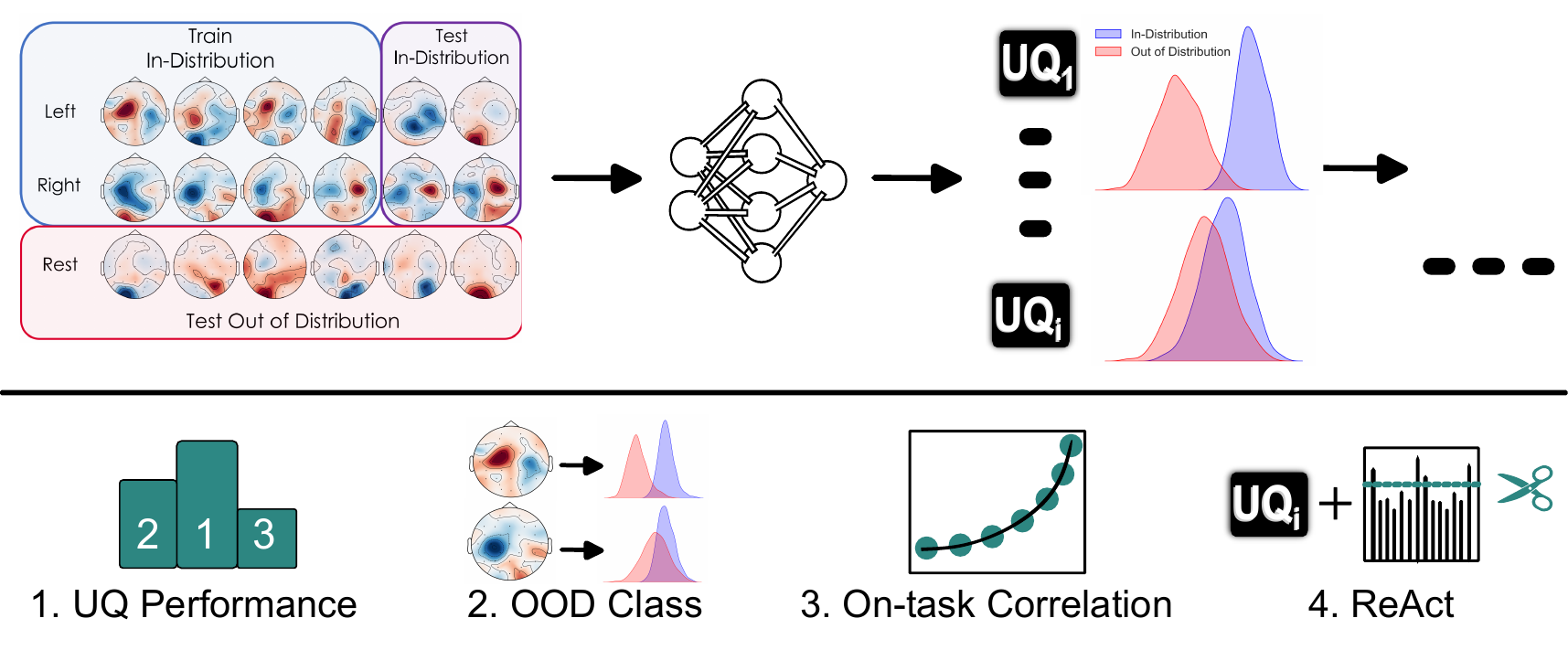}
    \caption{LOCO evaluation pipeline. \textit{Above the line:} We select a class as OOD and split the remaining data into ID train and test. A model is trained on the ID train data. After inference on the ID test and OOD data, we estimate the uncertainty scores for all tested uncertainty quantification (UQ) methods. The methods perform better with a smaller overlap in uncertainty scores between ID and OOD samples, as measured by AUROC. We repeat this procedure by selecting the next class as OOD until all classes have been selected as OOD once. \textit{Under the line:} We perform four distinct analyses on the gathered data using the pipeline described in section \ref{sec:analysis}. Within the first analysis, we compare the UQ methods across three different datasets and establish SOTA. Secondly, we establish differences in OOD detectability across motor imagery classes. Thirdly, we demonstrate a correlation between on-task and OOD detection performance. Lastly, we test whether OOD detection improves with ReAct.}
    \label{fig:LOCOEval}
\end{figure*}

\section{Methods}\label{sec:methods}

This paper performs a LOCO-OOD experiment, as introduced in Section \ref{sec:locoood}, for which we present the implementation below.
We then describe our CNN and the three datasets used in this experiment. Lastly, we discuss multiple analyses that we perform on the data gathered. Figure \ref{fig:LOCOEval} presents a full overview of the content of the methods sections.

\subsection{Experiment implementation}
We implemented a within-subject LOCO-OOD experiment as previously described in Algorithm \ref{alg:loco} and formally described in Appendix \ref{ap:Formalism}. We select a single class as OOD and train the model on the remaining classes. We rotate the selected OOD class. The model is not trained on the data of the OOD class. This represents real OOD detection conditions, where the user performs certain Motor Imagery tasks during a calibration session, but then might perform unfamiliar actions during inference, which do not belong to any of the known classes.

From the in-distribution classes, 10\% is used for validation and 15\% for testing. This was achieved using a stratified split. The total test set has a 1:1 ID-OOD ratio. The remaining 75\% of the ID dataset is used to train the feature extractor. 

We test seven different OOD detection methods spanning Bayesian Neural Network, distance-based, and density-based approaches as defined in Section \ref{sec:ood}, along with ReAct \cite{sun_react_2021} as an OOD optimisation method.
We run MC-Dropout with 50 forward passes and use 5 models in the Deep Ensemble.

The distance-based methods require hyperparameter tuning. We tune the hyperparameters on the ID testing data of the first user in the dataset and will use the found hyperparameters for the remaining users. The data of the first user is excluded during testing. For each combination of ID classes, we perform hyperparameter tuning without using any OOD data.

For d-KNN, for each \textit{k} we observe, for each ID testing data point, whether the \textit{k-th} nearest neighbour belongs to the same class. We choose the largest \textit{k} that classifies most data points correctly. We restrict the maximum \textit{k} to the mean number of training data points per class.

As DUQ has multiple hyperparameters that depend on each other, we use the Optuna hyperparameter optimisation framework \cite{optuna_2019} with the search space defined in Table \ref{tab:search_space}. We set the objective to find hyperparameters that yield the largest AUROC on the ID testing data. The identified optimal hyperparameters are available in Table \ref{tab:hyper_param_full}.

Furthermore, we train DUQ as a separate head with a learnable length scale per class on top of our frozen feature extractor. This differs from \cite{amersfoort_uncertainty_2020}, which jointly trains the DUQ head and the feature extractor. Moreover, our base model applies spectral normalisation only to the penultimate layer, instead of all constrained layers as suggested for DDU \cite{mukhoti_deep_2023}.   This provides a shared backbone between all OOD methods, but deviates from the original implementation. We provide results using the original implementations in Appendix \ref{ap:ablation}. Though we do not find statistically significant differences in OOD detection, our frozen feature extractor DUQ appears to outperform the original implementation in on-task classification on two datasets

We apply z-score normalisation to all EEG data after the split and before further usage.

\subsection{Base model}
As the base model for all experiments, we use a PyTorch implementation of EEGNeX \cite{chen_toward_2024}, a compact deep learning model designed for EEG signal classification. It consists of 8 convolutional blocks optimised for spatiotemporal feature learning across EEG channels and time windows.
Additionally, we extend the model with spectral normalisation applied to the penultimate convolutional linear layer for better class separation in latent space, which is essential for OOD methods such as DDU \cite{mukhoti_deep_2023}. Lastly, we train the model on batches of size 32 using a cross-entropy (CE) loss and an Adam optimiser \cite{adam} with a learning rate of $1 \cdot10^{-3}$ on all datasets. We use early stopping with a patience of 20 and half the learning rate with a patience of 5 to prevent overfitting, with a maximum of 200 epochs. The best-performing model from the training cycle is used in the experiments.

\subsection{Datasets}
In this study, we used three different MI-BCI and Motor Execution (ME) BCI datasets from MOABB \cite{MOABB}. The datasets complement each other as they differ in their technical setup, such as the number of channels, the sampling frequency, and the number of data points per user, as seen in Table \ref{tab:sum_dataset}. All datasets contain four classes corresponding to different movements, as seen in Table \ref{tab:movements}.
We used all the data after applying a bandpass filter from 8.0 to 32.0 Hz, following the default configuration in MOABB for the motor imagery paradigm. No ICA or artifact rejection is applied. Unless specified otherwise, we do not discard any data throughout the experiments.

\begin{table}
    \adjustbox{max width = \columnwidth}{
    \begin{tabular}{llll}
        \toprule
         & BNCI2014 &  Schirrmeister2017 & Stieger2021  \\
         \midrule
        \#Subjects & 9 & 12 & 62 \\
        \#Channels & 22 & 128 & 64 \\
        \#Classes & 4 & 4 & 4  \\
        \#Trials/class & 144 & 120 & 450 \\
        Trial length & 4s & 4s & 3s \\
        Frequency & 250Hz & 500Hz & 1000Hz \\ 
        \bottomrule
    \end{tabular}}
    \caption{Summary of the datasets.}
    \label{tab:sum_dataset}
\end{table}

\paragraph{BNCI2014} This is dataset IIA from the BCI competition IV \cite{BCNI2014}. The data is from a cue-based BCI paradigm corresponding to different MI tasks.
The participants were seated in a comfortable chair, watching a screen on which an arrow corresponding to a class was presented, prompting the participant to imagine the movement.

\paragraph{Schirrmeister2017} This dataset results from an experiment in which subjects perform left or right hand finger tapping, toe clenching or resting \cite{schirrmeister_deep_2017}. The movements were executed according to the direction of an arrow. The experiment was originally carried out in a lab setting with 128 channels, of which 44 were placed over the motor cortex. This dataset uses Motor Execution instead of Motor Imagery, but these exhibit similar evoked potentials and sensorimotor rhythms \cite{carrillo2008equivalent, kraeutner2014motor}. The types of movement performed should have negligible impact on head stabilisation and, therefore, EMG artifacts in the EEG recordings. 

\paragraph{Stieger2021} This dataset comes from a study investigating how individuals learn to control sensorimotor rhythm BCIs \cite{stieger2021}. In total, 33 participants went through an eight-week mindfulness program, and 29 participants were assigned to the control condition. Both groups received BCI training prior to the experimental data being obtained. During the experiment, participants had to move a cursor on a screen using hand movements. Data from both groups are used in this experiment.

\begin{table}[!t]
    \centering
    \adjustbox{max width = \columnwidth}{
    \begin{tabular}{ll}
    \toprule
    Class & Movement \\
    \midrule
    
    \addlinespace[0.8em]
    \multicolumn{2}{c}{\textit{BNCI2014}} \\
    \addlinespace[0.8em]
    
         Left hand & Movement left hand\\
         Right hand & Movement right hand\\
         Both feet & Movement feet \\
         Tongue & Movement tongue\\
         
    \addlinespace[0.8em]
    \multicolumn{2}{c}{\textit{Schirrmeister2017}} \\
    \addlinespace[0.8em]
    
        Left hand & Sequential finger tapping \\
        Right hand & Sequential finger tapping \\
        Feet & Clench toes \\
        Rest & Voluntary rest  \\

    \addlinespace[0.8em]
    \multicolumn{2}{c}{\textit{Stieger2021}} \\
    \addlinespace[0.8em]
    
        Left hand  & Left hand open and close \\
        Right hand & Right hand open and close \\
        Both hands & Both hands open and close \\
        Rest & Voluntary rest \\
    \bottomrule
    \end{tabular}}
    \caption{Classes and movements in the datasets. The Schirrmeister2017 dataset contains executed movements; the other datasets contain imagined movements.}    
    \label{tab:movements}
\end{table}

\subsection{Analyses}\label{sec:analysis}
We perform multiple analyses on the results of the predefined experiment. As seen in Figure \ref{fig:LOCOEval}, these include comparing OOD detection methods, differences between OOD classes, correlation with on-task performance, and an OOD optimisation technique.

\subsubsection{OOD approach comparison}
We perform the experiment on all three predefined datasets. For each dataset, we calculate the AUROC per OOD detection method for all participants and possible OOD classes. To explore the differences between the OOD detection methods and the datasets, we used a linear mixed-effects model (LMM). Within this model, we treat the participants as a random effect, as multiple data points belong to the same participant. 
The BCNI2014 dataset and Softmax will be the baseline to which interactions will be compared. 
Furthermore, we perform pairwise comparisons between the UQ methods using the estimated marginal means averaged over the datasets and a z-test with Holm correction. 
As a measure of central tendency, we report the median (Mdn) to account for the differing distributions of the AUROC per result, which do not always follow a normal distribution.

\subsubsection{Class dependent OOD detection}

As OOD performance might change depending on the class selected as OOD with different EEG features, we separate the AUROC values according to the OOD class and test significance using a Kruskal-Wallis test with Bonferroni corrected $\alpha{=}1.7 \cdot 10^{-2}$. Using a Dunn post hoc test, we determine which classes are significantly different.

\subsubsection{On-task classification vs OOD detection performance}

As better-performing models are more proficient at correctly classifying samples, it is likely that they can separate features in the latent space better. This has the potential to support OOD detection methods in distinguishing OOD from ID. Therefore, we test whether there is a correlation between the model's on-task performance and the OOD detection performance per detection approach using Spearman's Rank Correlation with Bonferroni corrected $\alpha{=}2.4 \cdot 10^{-3}$. As Deep Ensemble and DUQ involve additional trainable parameters, we evaluate the on-task performance of those additional parameters. 
As there is no meaningful difference in MI-BCI between MC Dropout classification and inference with a full model \cite{diffMCvsNormal}, we assume the same on-task performance.
We perform this analysis on both on-task AUROC gathered for models trained on three and two ID classes. 

\subsubsection{OOD optimization}

We determine whether ReAct \cite{sun_react_2021} can increase the ability of OOD detection approaches to distinguish OOD from ID. 
We keep the same pipeline as for the previous experiments. We add ReAct to the feature extractors' output before feeding it to the OOD detection methods and test this on all datasets. We determine significance using a Wilcoxon signed-rank test with Bonferroni corrected $\alpha{=}2.4 \cdot 10^{-3}$.

\section{Results}\label{sec:contents}

We show the differences in OOD-detection performance in both the methods and the datasets as the main analysis. Following this, we investigated class differences, the correlation between on-task performance and OOD detection, and optimisation methods as a further analysis.

\begin{figure}[!t]
    \centering

    \begin{subfigure}[b]{\linewidth}
        \includegraphics[width=1\linewidth]{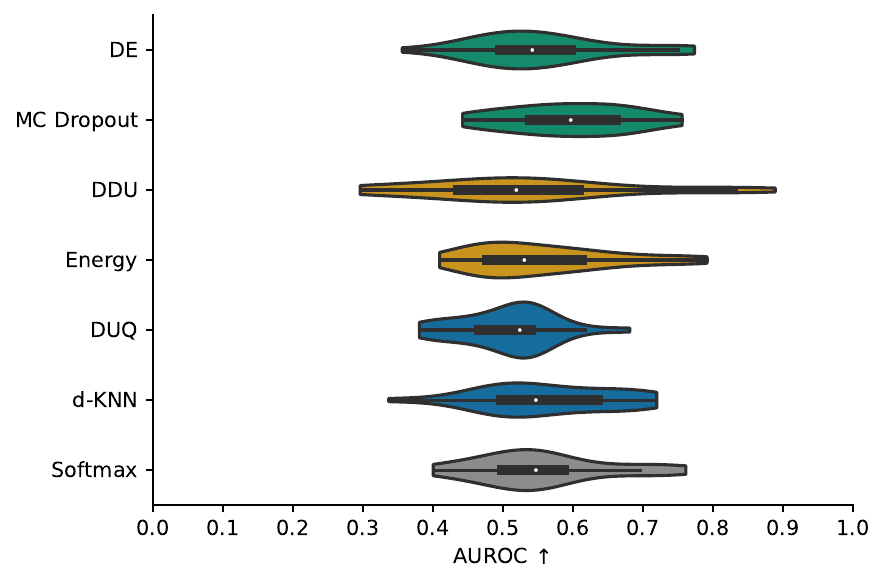}
        \caption{BNCI2014 dataset. MC Dropout performs best (Mdn=0.60, IQR=0.12).}
        \label{fig:results_BNCI2014}
    \end{subfigure}

    \begin{subfigure}[b]{\linewidth}
        \includegraphics[width=1\linewidth]{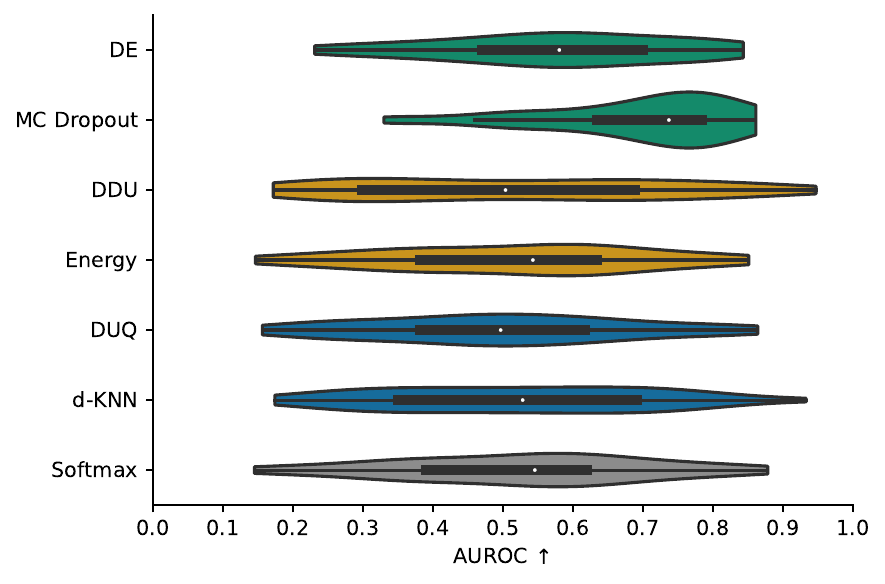}
        \caption{Schirrmeister2017 dataset. MC Dropout performs best (Mdn=0.74, IQR=0.15)}
        \label{fig:results_schirrmeister}
    \end{subfigure}

    \begin{subfigure}[b]{\linewidth}
        \includegraphics[width=1\linewidth]{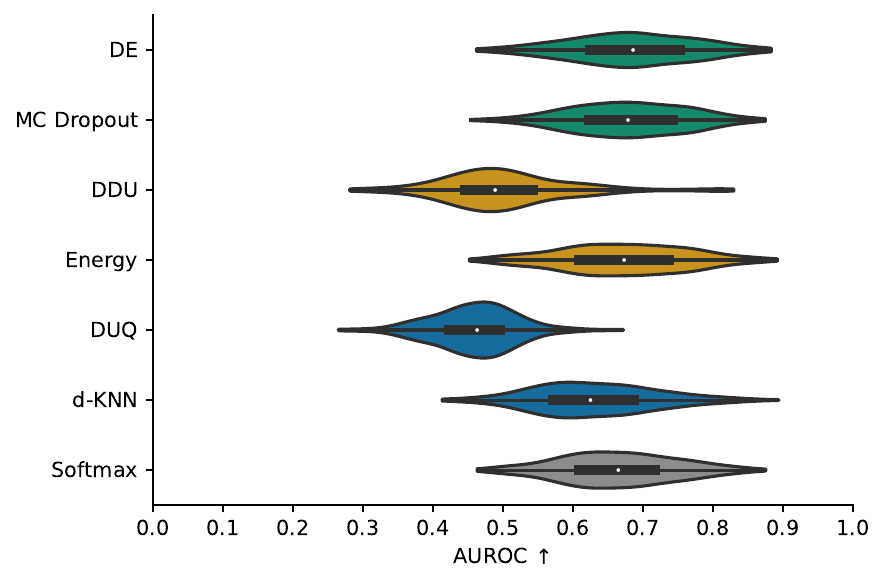}
        \caption{Stieger2021 dataset. Deep Ensemble (DE) performs best (Mdn=0.69, IQR=0.13).}
        \label{fig:stieger_comparison}
    \end{subfigure}

    \caption{Performance of the OOD detection methods in the LOCO-OOD experiment across three datasets. Colours indicate the following UQ categories:
    \protect\legendbox{bayes}{Bayesian},
    \protect\legendbox{density}{Density},
    \protect\legendbox{distance}{Distance},
    \protect\legendbox{baseline}{Baseline}. 
    MC Dropout separates ID and OOD samples best (all adjusted $p {<} 0.001$).}
    \label{fig:ood_comparison}
\end{figure}

\begin{figure}[!ht]
    \centering

    \begin{subfigure}[b]{\linewidth}
        \includegraphics[width=1\linewidth]{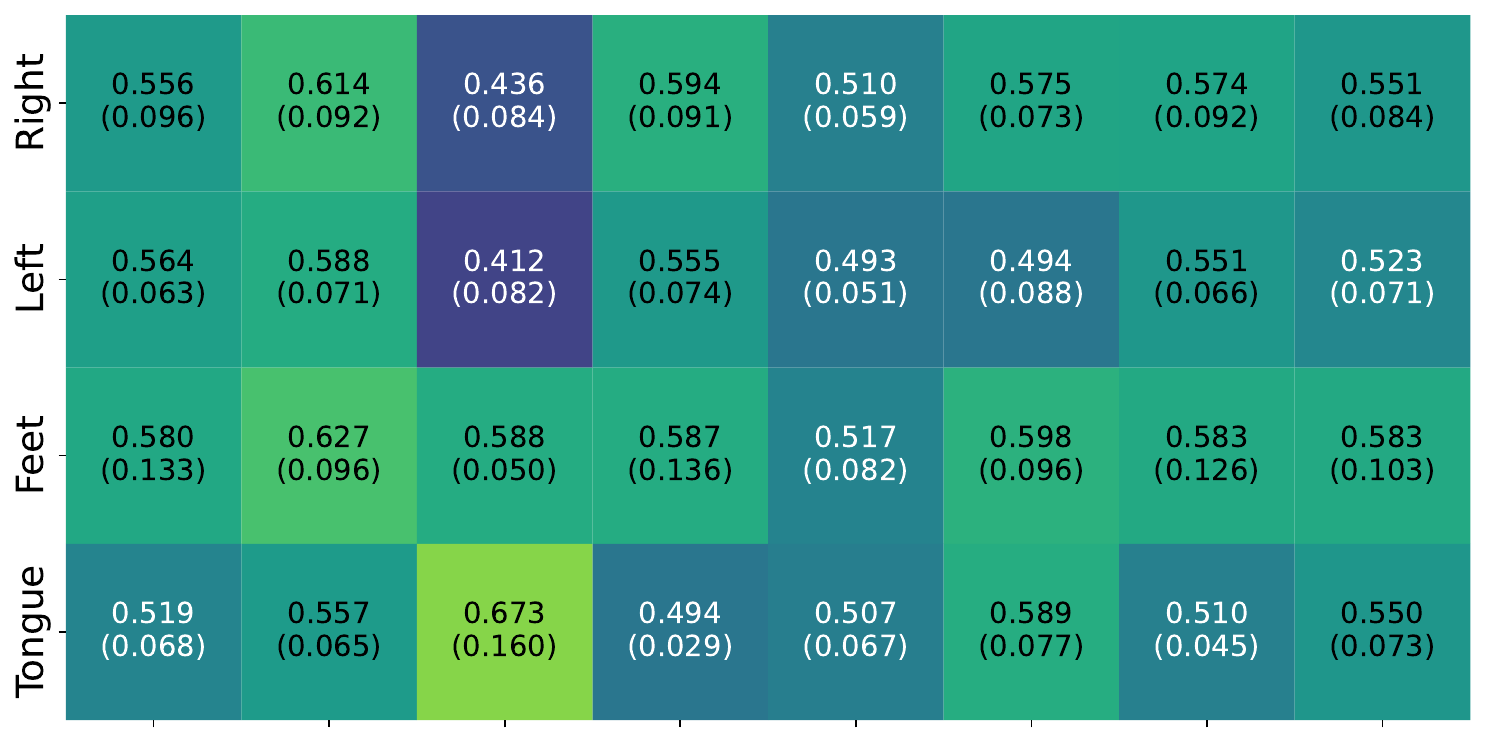}
        \caption{BNCI2014 dataset. Only minor differences between OOD classes.}
        \label{fig:BNCI_heatmap}
    \end{subfigure}

    \begin{subfigure}[b]{\linewidth}
        \includegraphics[width=1\linewidth]{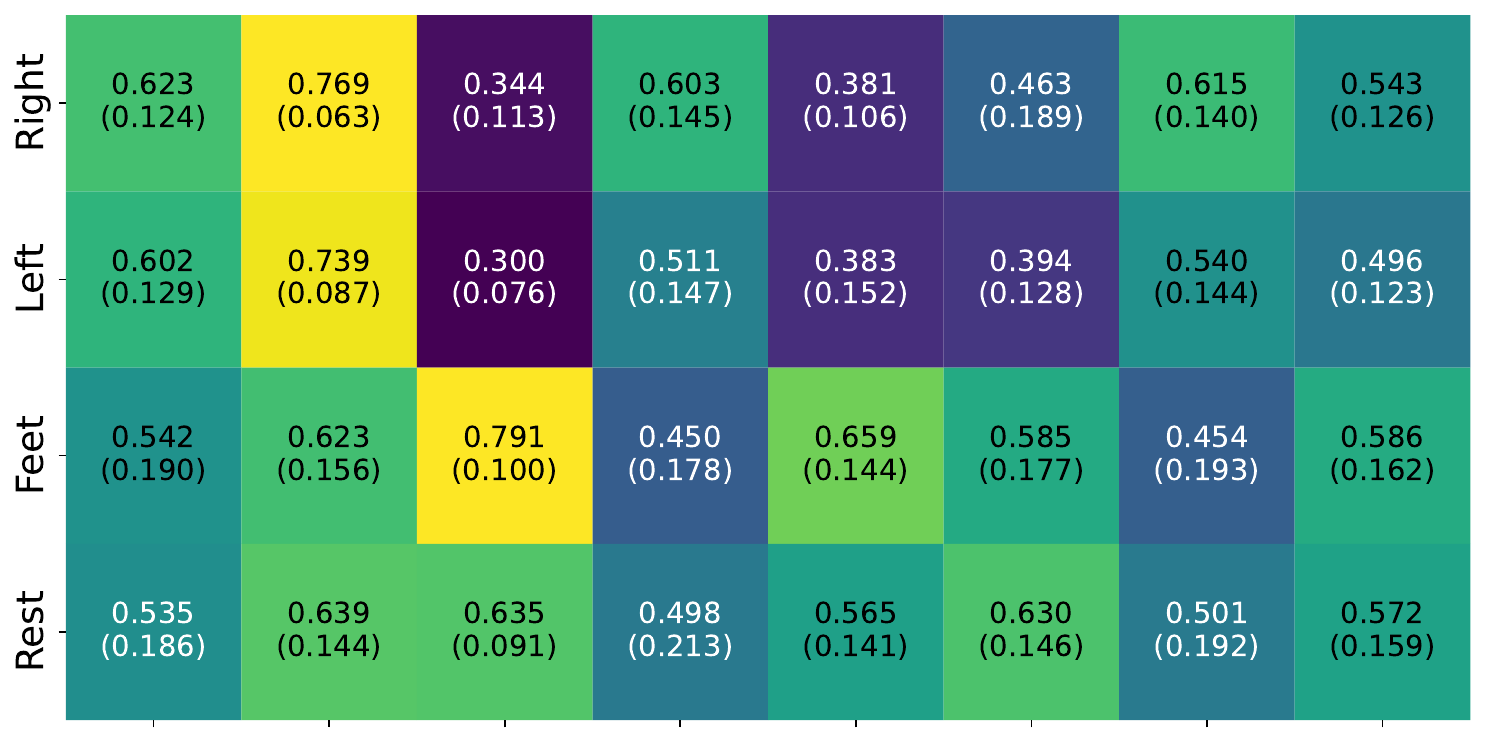}
        \caption{Schirrmeister2017 dataset. The Rest class is easier to OOD detect, although this varies between UQ methods.}
        \label{fig:Schirrmeister_heatmap}
    \end{subfigure}

    \begin{subfigure}[b]{\linewidth}
        \includegraphics[width=1\linewidth]{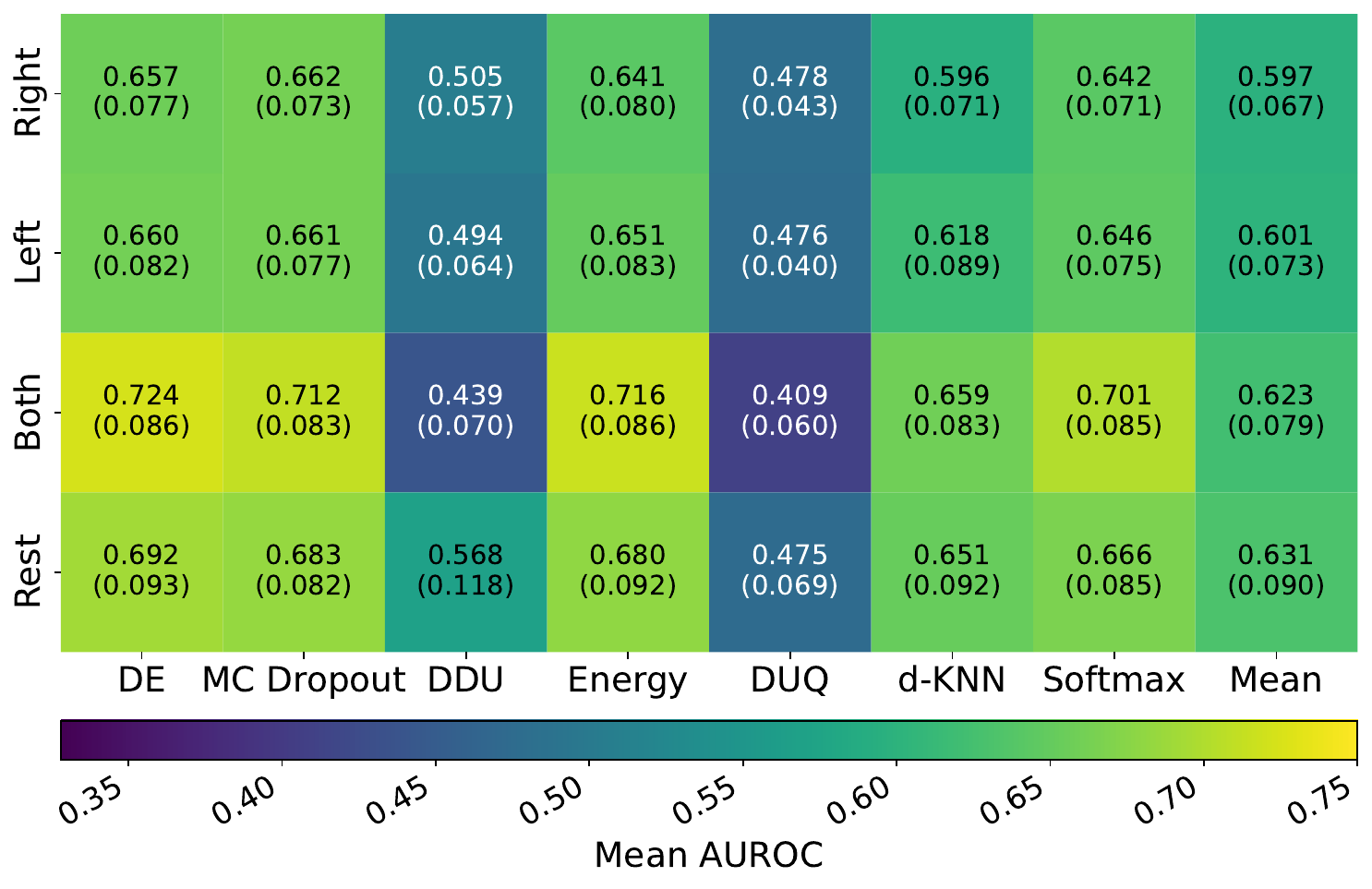}
        \caption{Stieger2021 dataset. The Both hands and Rest classes are easier to OOD detect.}
        \label{fig:Stieger_heatmap}
    \end{subfigure}

    \caption{Mean and standard deviation of OOD detection AUROC for different classes plotted against the different UQ methods. ``Left", ``Right" and ``Both" refer to hand movements. The last column represents the mean performance across methods. Whilst the performance for the Left and Right hand classes stays relatively similar, we find a significant increase in OOD detection performance for the Both hands and Rest classes.}
    \label{fig:class_difference}
\end{figure}

\subsection{OOD approach comparison}

MC Dropout receives the highest AUROC scores for OOD detection with MI-BCI data on both the BNCI2014 ($Mdn{=}0.60$, $IQR{=}0.12$) and Schirrmeister2017 ($Mdn{=}0.74$, $IQR{=}0.15$) datasets, as is evident in Figure \ref{fig:ood_comparison}. This is followed by Deep Ensembles, which receive the highest AUROC scores on the Stieger2021 dataset ($Mdn{=}0.69$, $IQR{=}0.13$).

We support this analysis with an LMM, which converged to an intercept estimate AUROC of $0.55$ for Softmax on the BNCI2014 dataset, which we define as the baseline.
Based on the estimated marginal means from the LMM across datasets, pairwise z-tests with Holm correction showed that MC Dropout outperforms all other methods (all adjusted $p {<} 0.001$).
DUQ and DDU perform significantly worse than the remaining methods (all adjusted  $p {<} 0.001$). d-KNN performed significantly worse than MC Dropout ($p{=} 1.7 \cdot 10^{-20}$, $\text{Estimated difference} = 0.09$) and Deep Ensembles ($p{=} 9.7 \cdot 10^{-4}$, $\text{Estimated difference} = 0.04$). We do not find a significant difference between the other approaches. The complete pairwise comparison is available in Table \ref{tab:pairwise}.

Furthermore, from the LMM we observe that DDU ($p{=}1.3 \cdot 10^{-8}$, $\text{Coef.}{=}-0.14$) and DUQ ($p{=}3.7 \cdot 10^{-11}$, $\text{Coef.}{=}-0.16$) show significantly worse performance compared to the other methods on the Stieger2021 dataset, as seen in Figure \ref{fig:stieger_comparison}. In contrast, generally, OOD detection methods perform better on the Stieger2021 dataset ($p{=}3.0 \cdot 10^{-5}$, $\text{Coef.}{=}0.11$). The full results of the LMM can be seen in Table \ref{tab:lmm_AUROC}. 

When models are trained on fewer classes, this can also affect OOD detection performance, as we demonstrate in Appendix \ref{ap:mltclasses}.

\begin{figure}[!t]
    \centering
    \includegraphics[width=\columnwidth]{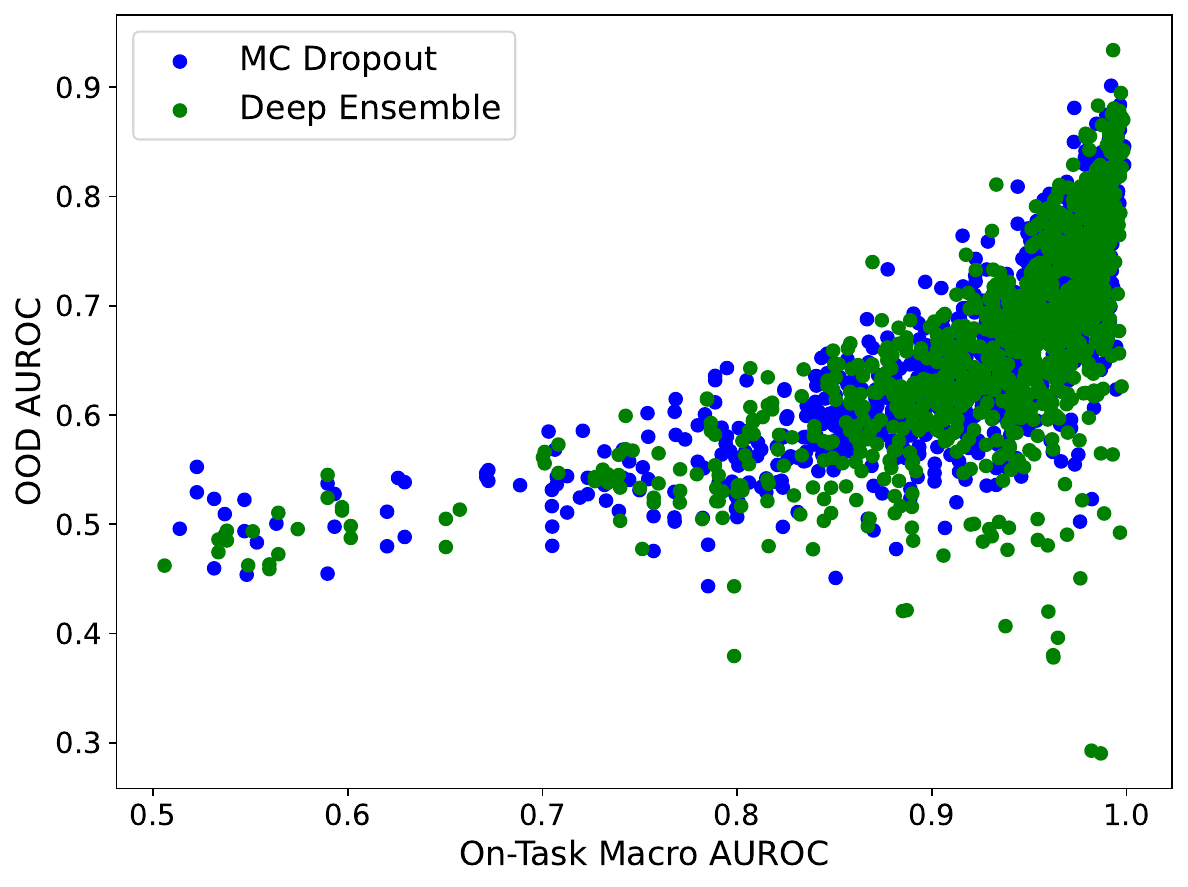}
    \caption{OOD detection vs on-task classification performance (macro-averaged across ID classes) on Bayesian Neural Network UQ approaches showing a non-linear correlation on the Stieger2021 dataset ($p{<}0.001$, $\rho{>}0.746$).}
    \label{fig:corr}
\end{figure}

\begin{table*}[!t]
    \centering
    \begin{tabular}{l c c cc c}
        \toprule
        \textbf{Method} & OOD-detection & On-Task & \multicolumn{2}{c}{Correlation} &  ReAct\\
        &  AUROC & AUROC &  $\rho$ & p-value & p-value \\
        \midrule
        \addlinespace[0.8em]
        \multicolumn{6}{c}{\textit{BNCI2014}} \\
        \addlinespace[0.8em]
        
        Deep Ensemble      & 0.542 (0.102) &  \textbf{0.852} & 0.586 & $\mathbf{3.88\cdot10^{-13}}$ & 0.026  \\
        MC Dropout         & \textbf{0.597 (0.124)} & 0.837  & 0.557 & $\mathbf{8.86\cdot10^{-12}}$ & 0.145 \\
        DDU                & 0.519 (0.174) & 0.837  & 0.052 & 0.558 &  0.945 \\
        Energy             & 0.530 (0.136) & 0.837 & 0.563 & $\mathbf{4.40\cdot10^{-12}}$ & 0.537 \\
        DUQ                & 0.524 (0.075) & 0.772 & 0.383 & 0.030 & 0.993 \\
        d-KNN                & 0.547 (0.139) & 0.837  & 0.504 & $\mathbf{2.24 \cdot10^{-3}}$ & 0.182 \\
        Softmax            & 0.547 (0.089) & 0.837 & 0.584 & $\mathbf{4.86\cdot10^{-13}}$ & 0.737 \\
 
        \addlinespace[0.8em]
        \multicolumn{6}{c}{\textit{Schirrmeister2017}} \\
        \addlinespace[0.8em]
        
        Deep Ensemble    & 0.580 (0.231) & \textbf{0.961} & 0.380 & $\mathbf{1.45\cdot10^{-8}}$ & 0.639 \\
        MC Dropout       & \textbf{0.737 (0.150)} & 0.958 & 0.425 & $\mathbf{1.53\cdot10^{-10}}$ & 0.171 \\
        DDU              & 0.504 (0.391) & 0.958 & 0.077 & 0.271  & 0.698 \\
        Energy           & 0.543 (0.255) & 0.958  & 0.353 & $\mathbf{1.76\cdot10^{-7}}$ & 0.013 \\
        DUQ              & 0.497 (0.237) & 0.816 & -0.225 & 0.109 & 0.940 \\
        d-KNN              & 0.528 (0.343) & 0.958 & -0.035 & 0.803 & 0.761 \\
        Softmax          & 0.545 (0.230) & 0.958 & 0.367 & $\mathbf{4.84\cdot10^{-8}}$ & $8.73\cdot10^{-3}$ \\

        \addlinespace[0.8em]

      \multicolumn{6}{c}{\textit{Stieger2021}} \\
        \addlinespace[0.8em]
        
        Deep Ensemble   & \textbf{0.686 (0.129)} & \textbf{0.923} & 0.746 & $\mathbf{4.30\cdot10^{-174}}$ &  1.000 \\
        MC Dropout      & 0.679 (0.121) & 0.912 & 0.830 & $\mathbf{3.14\cdot10^{-238}}$ &  1.000 \\
        DDU             & 0.489 (0.100) & 0.912 & -0.053 & 0.099 &  $\mathbf{1.03\cdot10^{-12}}$ \\
        Energy          & 0.673 (0.130) & 0.912  & 0.755 & $\mathbf{1.36\cdot10^{-180}}$ &  1.000\\
        DUQ             & 0.463 (0.074) & 0.594 & -0.091 & 0.154 & 0.619 \\
        d-KNN             & 0.625 (0.116) & 0.912 & 0.679  & $\mathbf{2.40 \cdot10^{-34}}$ &  1.000 \\
        Softmax         & 0.665 (0.109) & 0.912 & 0.771 & $\mathbf{1.89\cdot10^{-193}}$ & 1.000 \\
       
        \addlinespace[0.8em]
        \bottomrule
    \end{tabular}
\caption{Main results of the LOCO-OOD experiments. The OOD-detection AUROC is reported as the median with interquartile range. On-Task AUROC refers to the average classification performance on ID data for a model trained on three ID classes.
Using Spearman's rank correlation test between classification performance and OOD detection performance, reported as $\rho \ [\text{p-value]}$, we find a non-linear correlation across multiple results. 
We find that ReAct does not improve OOD optimisation following a Wilcoxon signed-rank test, for which we report the p-value. Boldface indicates the best AUROC value within each dataset and statistically significant p-values according to a Bonferroni corrected significance threshold $\mathbf{\alpha{=}2.4 \cdot 10^{-3}}$ to account for multiple testing. Note that the participants in the Schirrmeister2017 dataset performed Motor Execution, and participants in the Stieger2021 dataset participated in extensive training, explaining their high on-task AUROC.}
\label{tab:results_overview}
\end{table*}

\subsection{Class-dependent OOD detection}

Since different motor imagery classes correspond to distinct neural and muscular activity patterns, OOD detectability may not be uniform across classes.
When comparing AUROC scores among different OOD classes within each dataset, we observe class-dependent differences in OOD detectability as shown in Figure \ref{fig:class_difference}. A Kruskal–Wallis test with Bonferroni corrected significance threshold $\alpha{=}1.666 \cdot 10^{-2}$ confirms that these differences are statistically significant for the Schirrmeister2017 ($H = 11.9$, $p = 7.7 \cdot 10^{-3}$) and Stieger2021 ($H = 27.9$, $p = 3.9 \cdot 10^{-6}$) datasets, but not for BNCI2014 ($H = 8.0$, $p = 4.5 \cdot 10^{-2}$).

Post-hoc pairwise comparisons using Dunn`s test on the Schirrmeister2017 dataset reveal that the Rest class is easier to detect as OOD than the Left hand class ($p {=} 8.1\cdot 10^{-3}$), as shown in Figure \ref{fig:Schirrmeister_heatmap}. However, this class-dependent behaviour is not consistent across all detection methods. For example, hand movement classes achieve higher AUROC scores when using MC dropout, whereas the opposite trend is observed when using DDU.

Applying the same post-hoc analysis to the Stieger2021 dataset, we find that the Rest class is significantly easier to detect as OOD than both the Right hand ($p {=} 1.4 \cdot 10^{-4}$) and Left hand ($p {=} 2.2 \cdot 10^{-3}$) classes. 
As shown in Figure \ref{fig:Stieger_heatmap}, a similar effect is observed for the Both hands class, which is easier to detect as OOD than the Right ($p {=} 3.0 \cdot 10^{-4}$) and Left hand ($p {=} 4.0 \cdot 10^{-3}$) classes.

These results suggest that some individual OOD classes are more distinct from the ID data than others. When a detector is evaluated only on these easily separable classes and achieves a high average AUROC, this can be misleading. Such a detector may still fail to reliably flag other, less distinct off-task movements in practice.

\begin{figure}[!t]
    \centering
    \includegraphics[width=1\linewidth]{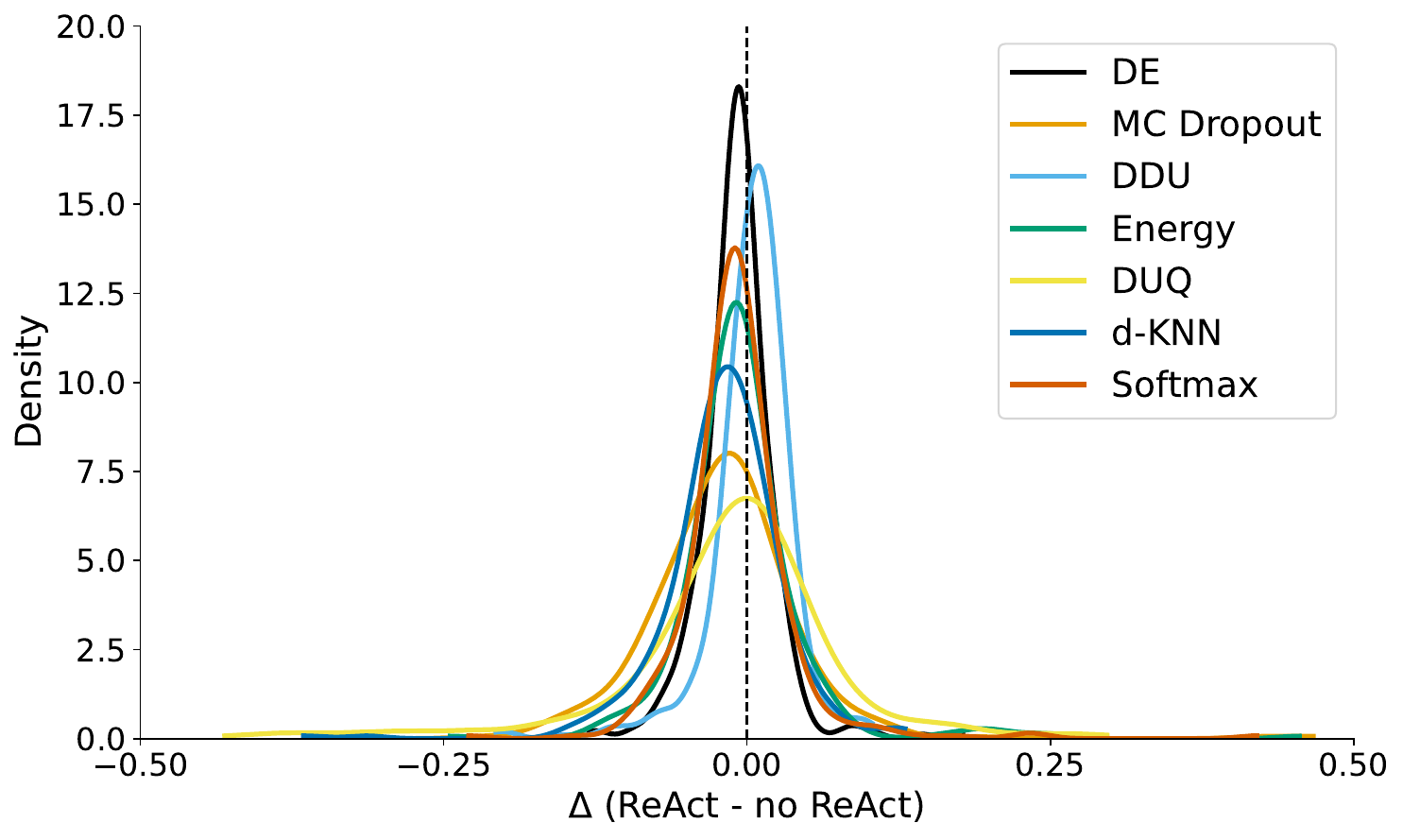}
    \caption{Difference in AUROC when using ReAct aggregated over all datasets. Smoothing was done using KDE. The dotted line indicates no difference. Whilst DDU experiences a minor benefit from ReAct, the other methods do not.}
    \label{fig:react}

\end{figure}

\subsection{On-task classification vs OOD detection performance}

We examine the relationship between classification performance and OOD detection performance. We see a non-linear relation in Figure \ref{fig:corr}, where the OOD detection performance increases with the classification performance, with further results given in Appendix \ref{ap:corr}. Table \ref{tab:results_overview} shows that this correlation exists for all OOD detection methods except DDU and DUQ. We can conclude, for all other methods on all the tested datasets, that it is easier to detect an OOD sample with high on-task performance, suggesting that as models and users get better, OOD detection performance will also increase.

\subsection{OOD optimisation}
OOD detection methods can benefit from ReAct as an optimisation. We analysed whether ReAct is effective for OOD detection with MI-BCI as well. 

Figure \ref{fig:react} illustrates how ReAct affects performance when activations are clamped at the 90th percentile of the ID train data. Accounting for multiple comparisons with Bonferroni correction in Table \ref{tab:results_overview}, ReAct does not show a significant improvement over the other models and datasets, except for DDU ($p{=}1.0\cdot10^{-12}$), where we observe that the peak lies slightly after 0, indicating a small improvement.
Overall, ReAct does not give a substantial and consistent improvement in OOD detection.

\section{Discussion}\label{sec:conclusions}

Well-performing OOD detection in MI-BCI has the potential to make BCI systems safer when users show off-task behaviour. As OOD detection is an underexplored topic in Brain-Computer Interfaces, our objective was to benchmark OOD detection methods in MI datasets using CNN-based architectures and to identify possible reasons for their performance. In the following sections, we describe our main findings and interpretations, limitations, and recommendations for future research.

\subsection{Findings}

Benchmarking studies such as \citet{mucsanyi_benchmarking_2024} performed with image data show not only better OOD detection but also a smaller variance between models and runs than we find.
BCI data is more difficult to classify due to noisy signals and uncertainties in the data. Given the differences in data, the findings from the computer vision domain are not immediately transferable to BCI applications, as illustrated by our findings.

We conclude that MC Dropout is the most reliable OOD detection method for Motor Imagery BCIs, followed by Deep Ensembles. Deep Ensembles and MC Dropout are both discriminative methods and approximations of Bayesian Neural Networks. This suggests that previous studies on Bayesian Neural Networks in BCIs \cite{UNCER, Kim2026, Milanes2021, BayesianNetworksBCI, suurmeijer2025uncertainty,} are also promising for OOD detection. The Deep Ensemble also yield strong on-task performance. While OOD detection performance was not good on all datasets, we find that when on-task performance is high, OOD detection is effective, contradicting previous null-results from \citep{jong_how_2024}.

DDU and DUQ yielded less reliable OOD detection performance across datasets, consistent with previous findings for DUQ in similar settings \cite{suurmeijer2025uncertainty, diffMCvsNormal}. DDU and DUQ differ from the other detection methods as they can be viewed as generative methods \cite{diffMCvsNormal} that rely on explicitly learned structures in the latent space. Due to the low signal-to-noise ratio inherent to EEG data and the presence of label noise, modelling the ID distribution in the latent space becomes increasingly difficult, causing ID samples to receive high uncertainty estimates and ultimately reducing OOD detection performance.

We found substantial differences in the difficulty of OOD detection between different datasets.
In particular, the Stieger2021 dataset contained more samples per participant per class, providing a better approximation of the distribution of possible ID samples. As a machine learning model can improve by being trained on more data, this can lead to a better model and possibly more reliable OOD detection. 

Moreover, we found that some OOD classes are easier to detect than others and found a correlation between the on-task model performance and the OOD detection performance, especially for the Bayesian Neural Network and density-based methods. This reinforces that in conditions where on-task performance is high, as is required for deployment, OOD detection performance will also be high. This yields promising results for implementing OOD detection in practical BCI use to improve safety against off-task behaviour.

\citet{sun_react_2021} motivated ReAct on the argument that OOD samples can have higher activations than ID samples, leading to overconfidence, and, hence, activations should be clamped. However, we find that, on average, ReAct does not improve the OOD detection methods for Motor Imagery BCIs. This shows that not all existing methods are effective.

Within the experiment, we noticed that methods in some runs performed worse than random. 
This means that the UQ method assigns higher uncertainty to ID samples, while that should be the case for OOD samples.

\subsection{Limitations \& Future Research}

While we found a strong correlation between on-task performance and OOD detection, it is not perfect. While users with good on-task performance generally have good OOD detection performance, this is not always the case. We found some users who have excellent on-task performance (AUROC $\>$ 0.95), but below random-guessing OOD detection performance. It is not clear what might cause this, and it means that solely good on-task performance is not sufficient to conclude that OOD detection is reliable. Effectively, this means that for usage, OOD detection performance still needs to be established separately using the proposed experimental setup. Future research may attempt to identify whether these users get consistently worse OOD detection performance, and if so, identify what might be causing that.

Our experiments rely on a LOCO-OOD task. This means that we test the OOD detection ability methods on classes unknown to the model. While this is one of the challenges for a robust BCI system, there are other OOD applications for BCIs. 
\citet{EEGShifts} describes several alternative EEG shifts that may be considered as alternative ecologically valid types of OOD data. These include shuffled channels, amplitude modulation, slow drift and time warping. In these cases, it is harder to determine whether a model should make correct predictions despite the shifts or whether it should conclude that it is not possible to make a prediction. 

All evaluated methods operate on CNN-based EEG representations, which may implicitly capture spectral and spatial structure through learned filters. In contrast, traditional methods such as CSP-LDA or MDRM explicitly exploit the physiological characteristics of EEG, yielding more interpretable representations, but remain underexplored for OOD detection despite providing good uncertainty estimates \cite{suurmeijer2025uncertainty}. Future work could compare these approaches for OOD detection in terms of performance and interpretability.

Additionally, OOD detection may be explored in other BCI modalities, particularly in invasive BCIs. Studies with implanted BCIs in ALS patients reported difficulty during the night, which was previously addressed with a dedicated nightmode functionality \citep{leinders2026implanted}, but may also be treated as an OOD condition. In implanted BCIs, the signal-to-noise ratio is typically more favourable, which is promising for better OOD detection.

\subsection{Conclusion}
In summary, we found several UQ methods that perform well in OOD detection. We recommend MC Dropout and Deep Ensembles for their good on-task performance and OOD detection performance.

The correlation between on-task performance and OOD detection performance is promising, as it shows that conditions where BCIs are best in-distribution are also best for OOD detection. We found OOD detection performance with an AUROC up to 0.7, suggesting that these methods can work. However, there is a large inter-subject variability that is exhibited in both on-task performance and OOD detection performance, limiting the universality of OOD detection in BCIs.

\data{The codebase for this study is available at the following GitHub repository: \url{https://github.com/MerlijnQ/OOD-BCI}}

\renewcommand*{\bibsetup}{\raggedright}
\printbibliography

\begin{appendices}

\renewcommand{\thefigure}{\thesection\arabic{figure}}
\renewcommand{\thetable}{\thesection\arabic{table}}
\makeatletter
\@addtoreset{figure}{section}
\@addtoreset{table}{section}
\makeatother

\clearpage
\onecolumn  
\section{Classification vs OOD detection performance}\label{ap:corr}

\begin{figure*}[htp]

    \centering
    \begin{subfigure}[t]{0.3\textwidth}
        \centering
        \includegraphics[width=\linewidth]{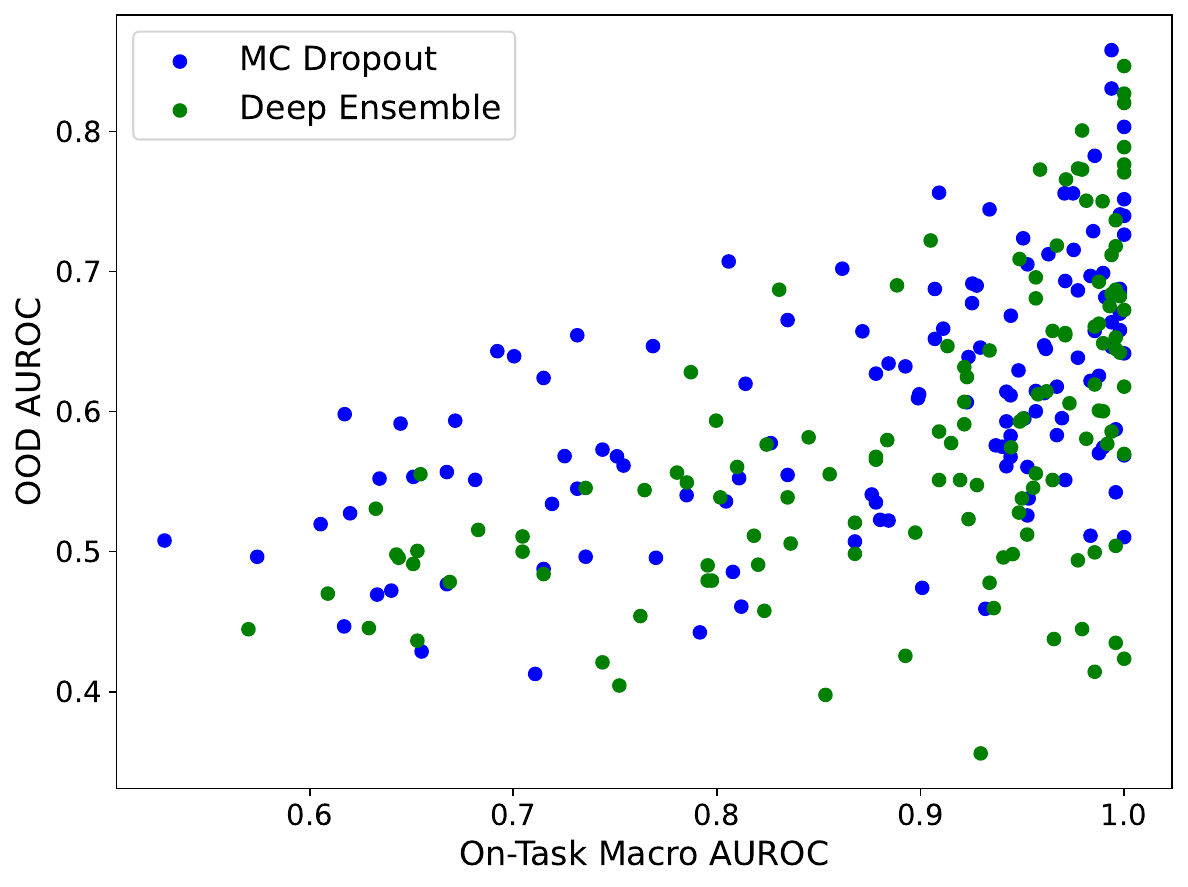}
        \caption{Bayesian - BNCI2014}
    \end{subfigure}
    \hfill
    \begin{subfigure}[t]{0.3\textwidth}
        \centering
        \includegraphics[width=\linewidth]{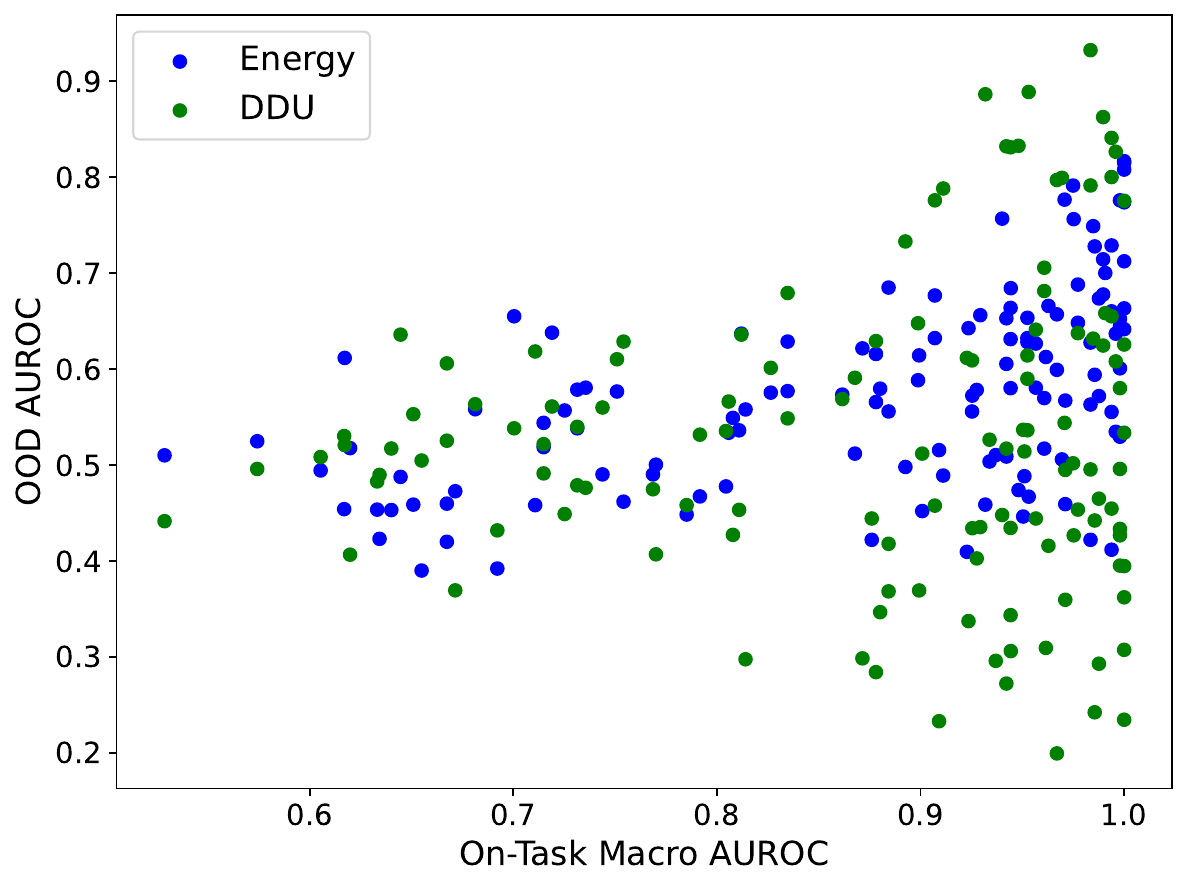}
        \caption{Density - BNCI2014}
    \end{subfigure}
    \hfill
    \begin{subfigure}[t]{0.3\textwidth}
        \centering
        \includegraphics[width=\linewidth]{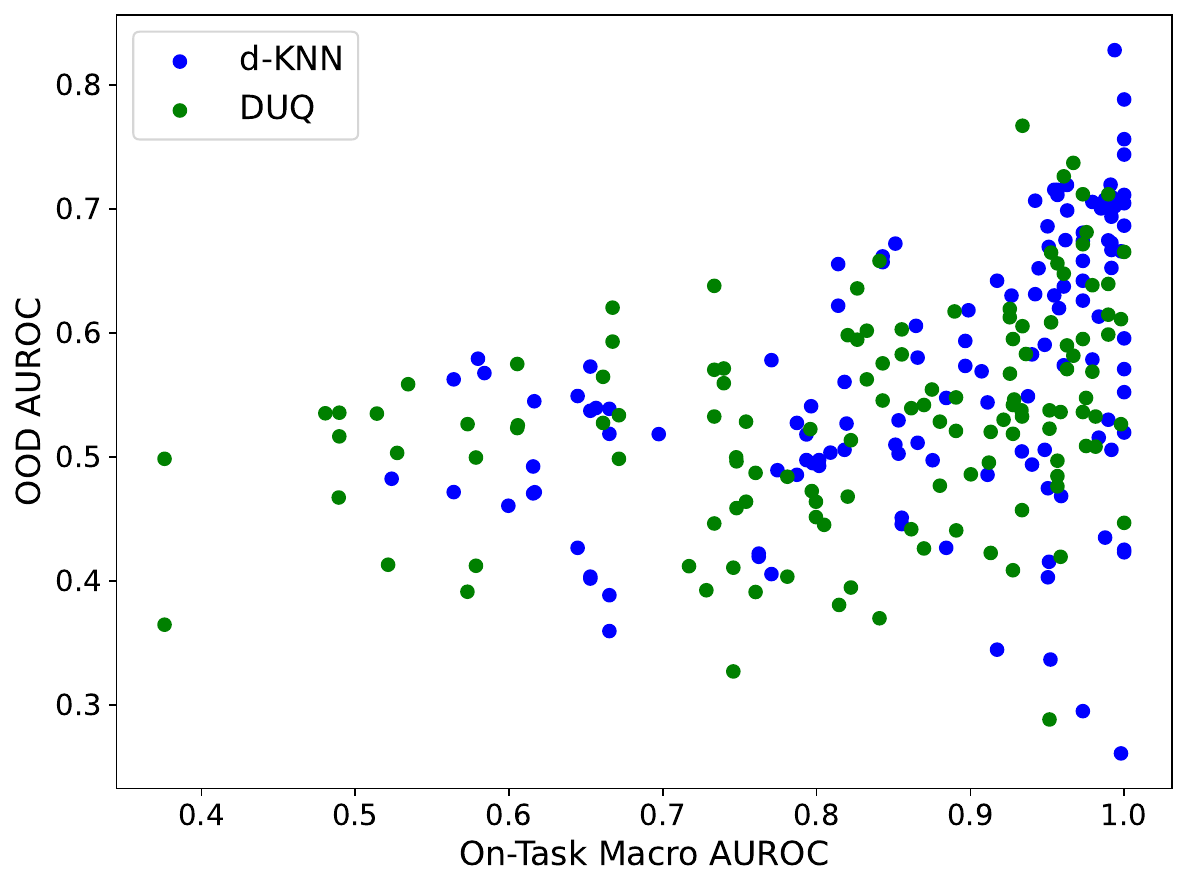}
        \caption{Distance - BNCI2014}
    \end{subfigure}
    
    \vspace{1em}

    \begin{subfigure}[t]{0.3\textwidth}
        \centering
        \includegraphics[width=\linewidth]{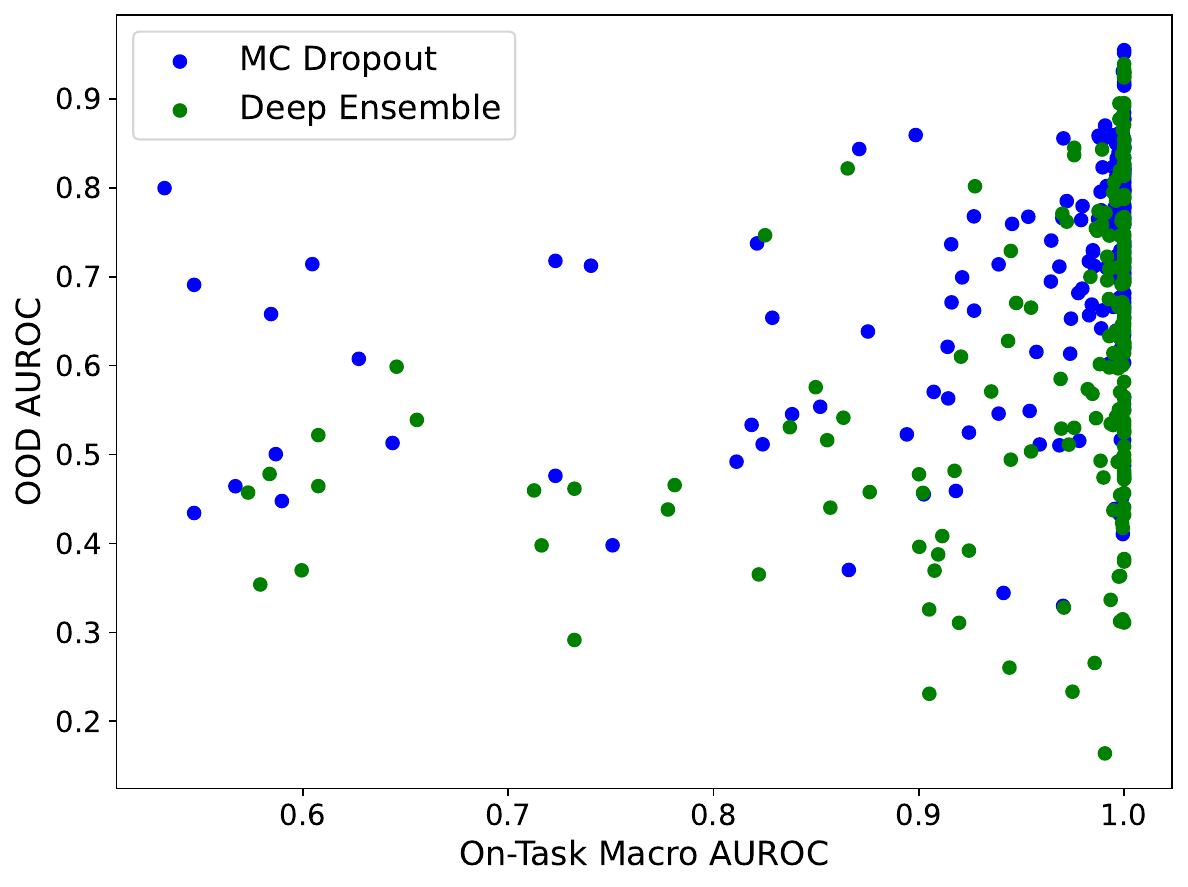}
        \caption{Bayesian - Schirrmeister2017}
    \end{subfigure}
    \hfill
    \begin{subfigure}[t]{0.3\textwidth}
        \centering
        \includegraphics[width=\linewidth]{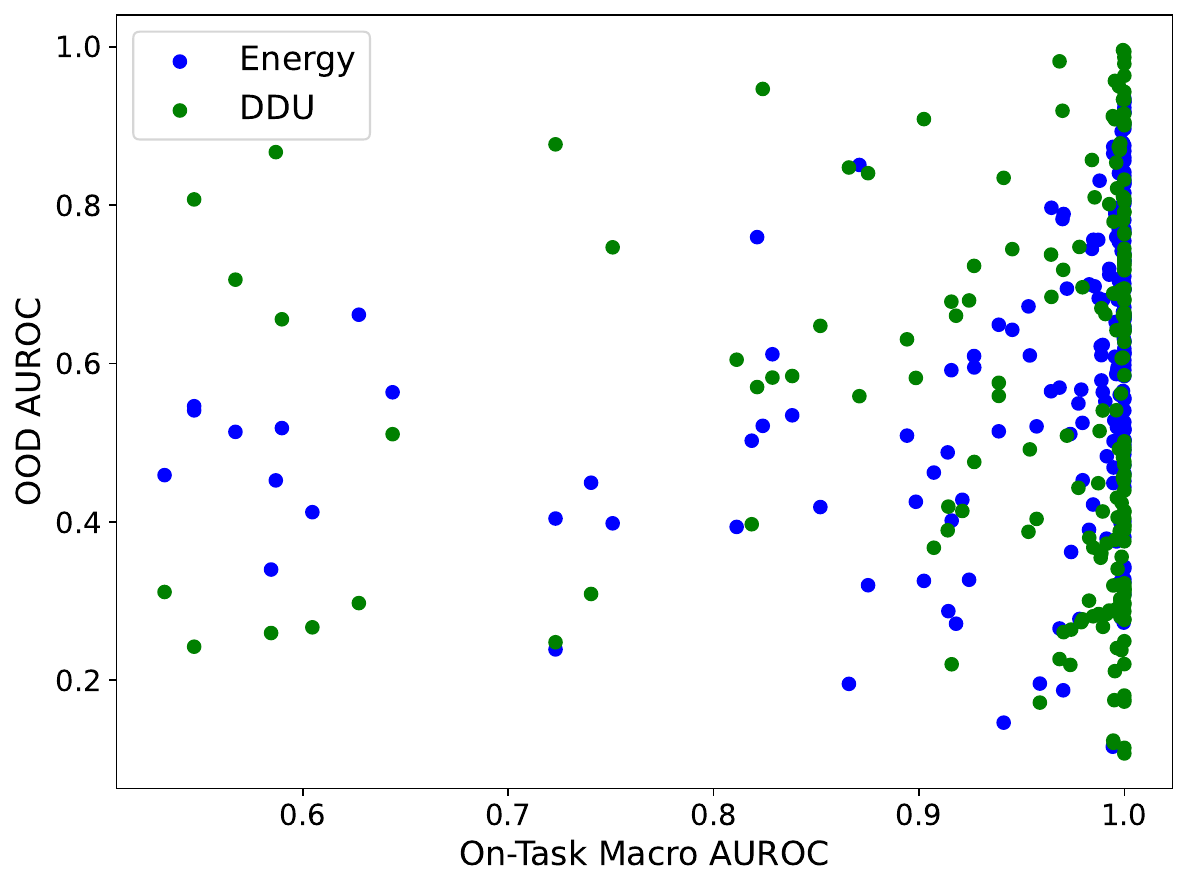}
        \caption{Density - Schirrmeister2017}
    \end{subfigure}
    \hfill
    \begin{subfigure}[t]{0.3\textwidth}
        \centering
        \includegraphics[width=\linewidth]{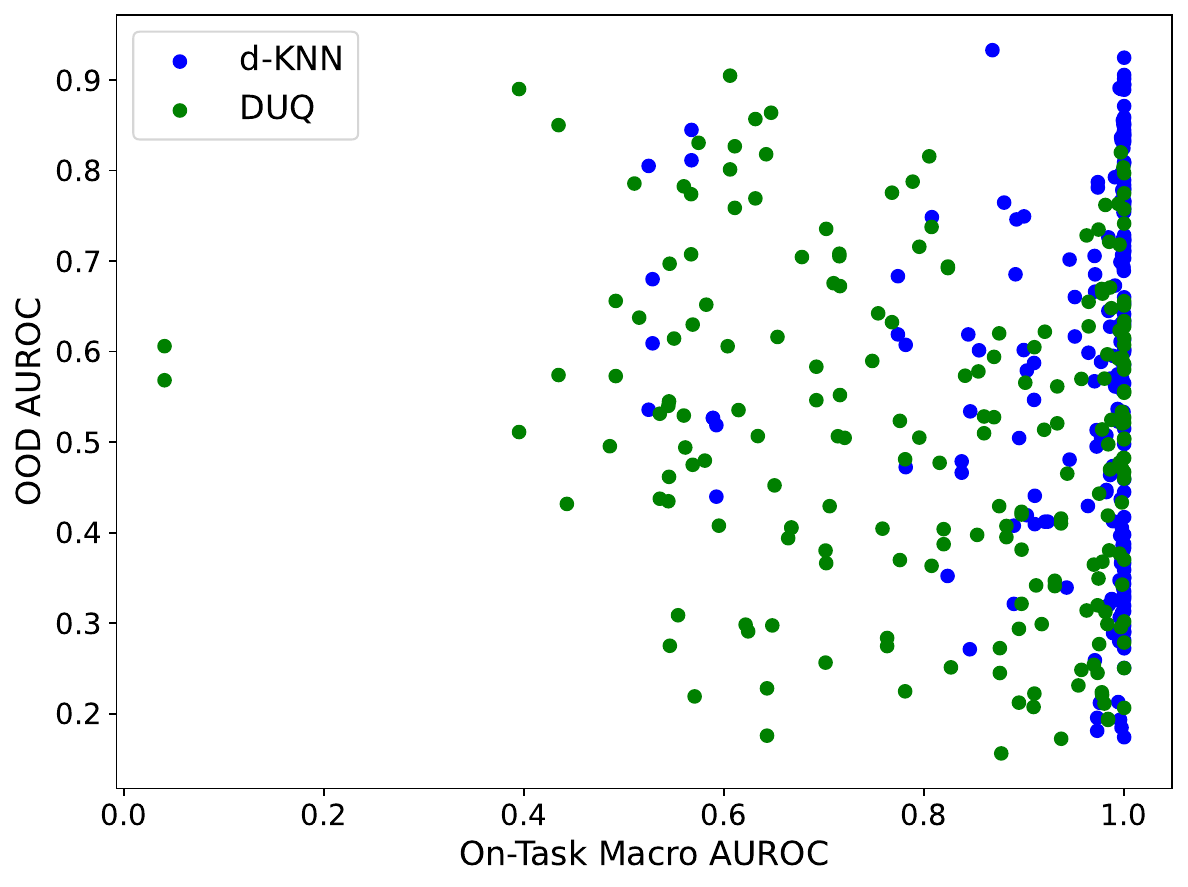}
        \caption{Distance - Schirrmeister2017}
    \end{subfigure}

    \vspace{1em}

    \begin{subfigure}[t]{0.3\textwidth}
            \centering
            \includegraphics[width=\linewidth]{img/correlation/Stieger2021_results_MC_Dropout_Deep_Ensemble.pdf}
            \caption{Bayesian - Stieger2021}
        \end{subfigure}
        \hfill
        \begin{subfigure}[t]{0.3\textwidth}
            \centering
            \includegraphics[width=\linewidth]{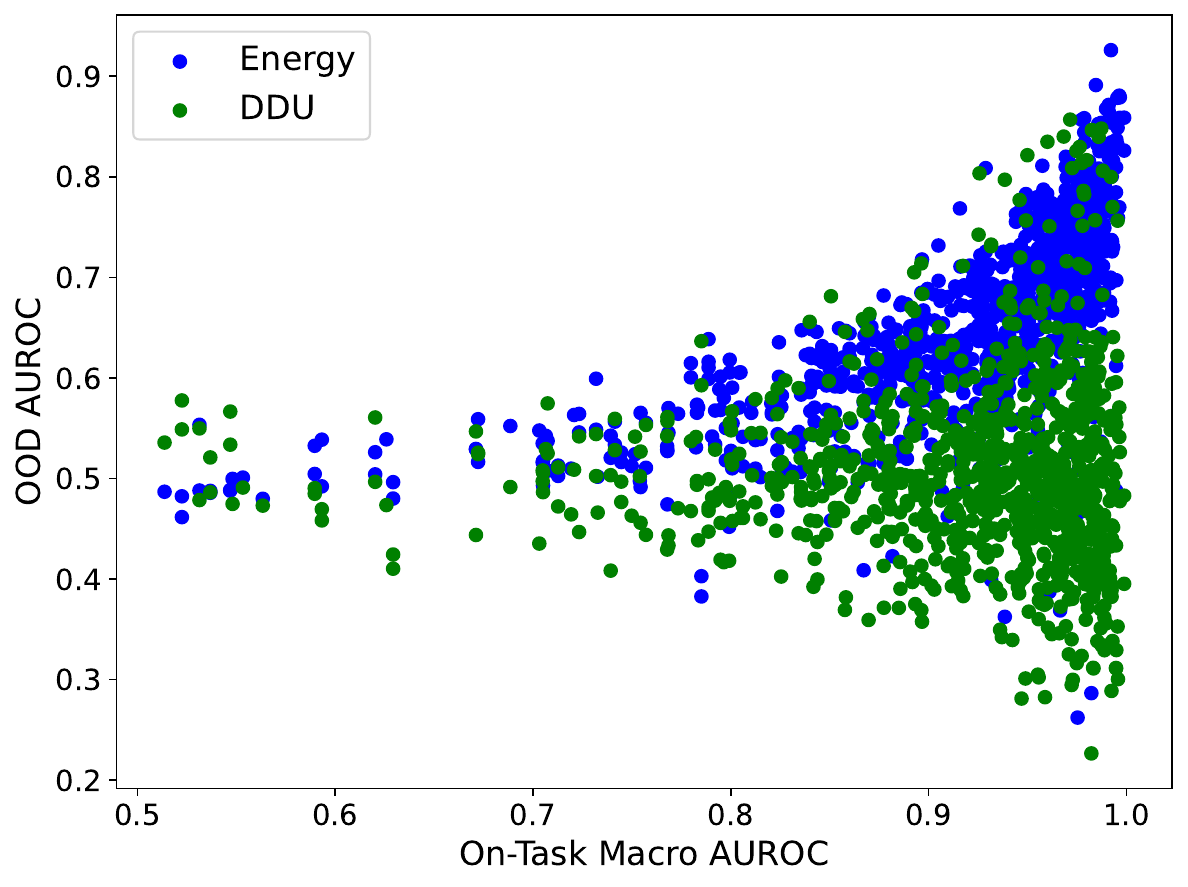}
            \caption{Density - Stieger2021}
        \end{subfigure}
        \hfill
        \begin{subfigure}[t]{0.3\textwidth}
            \centering
            \includegraphics[width=\linewidth]{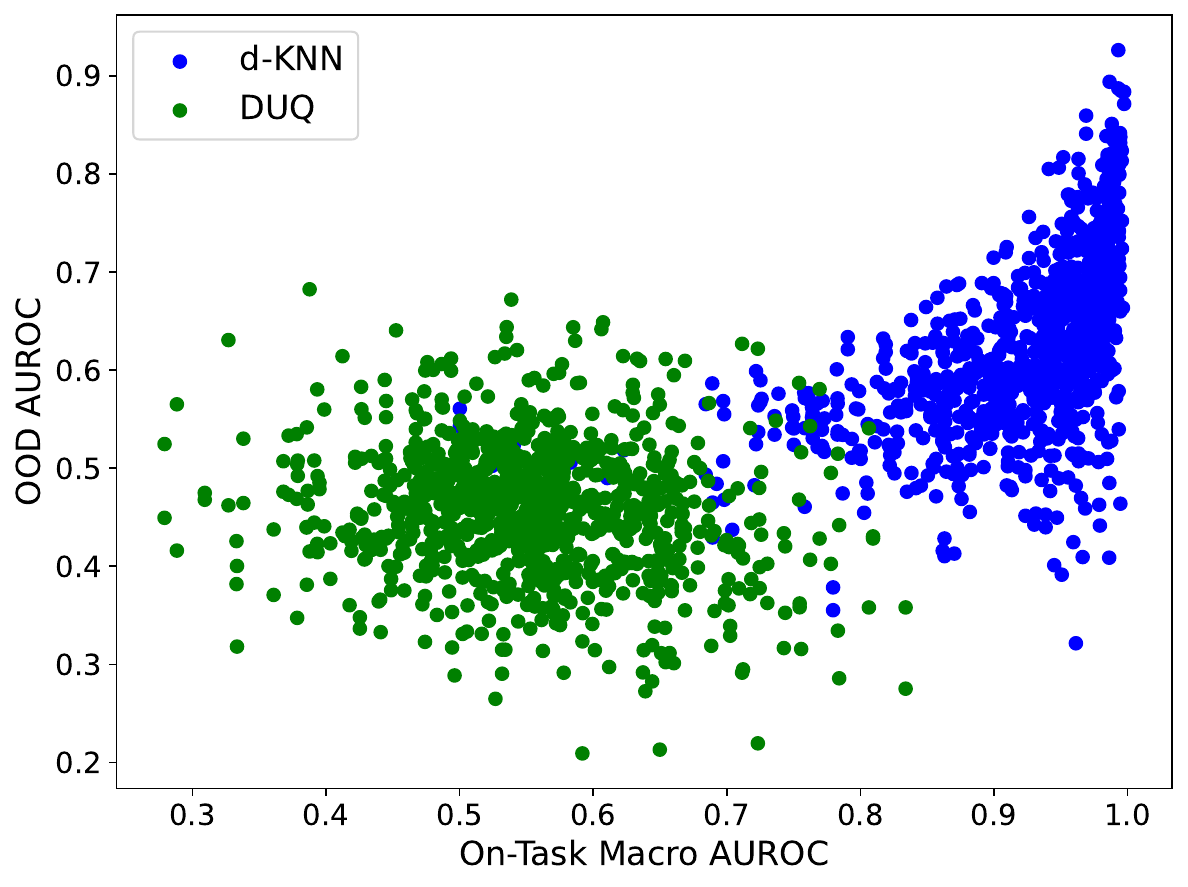}
            \caption{Distance - Stieger2021}
        \end{subfigure}

    \caption{OOD detection vs on-task classification performance.}
    \label{fig:compare_corralations}
\end{figure*}

Figure \ref{fig:compare_corralations} demonstrates the non-linear correlation between on-task performance and OOD detectability per dataset and UQ category. This correlation indicates that on-task performance is predictive of OOD detection performance with the exception of DDU and DUQ.

\clearpage

\section{Difference in number of ID classes}\label{ap:mltclasses}

\begin{figure}[h!]
    \centering
    \begin{subfigure}[t]{0.32\textwidth}
        \includegraphics[width=\linewidth]{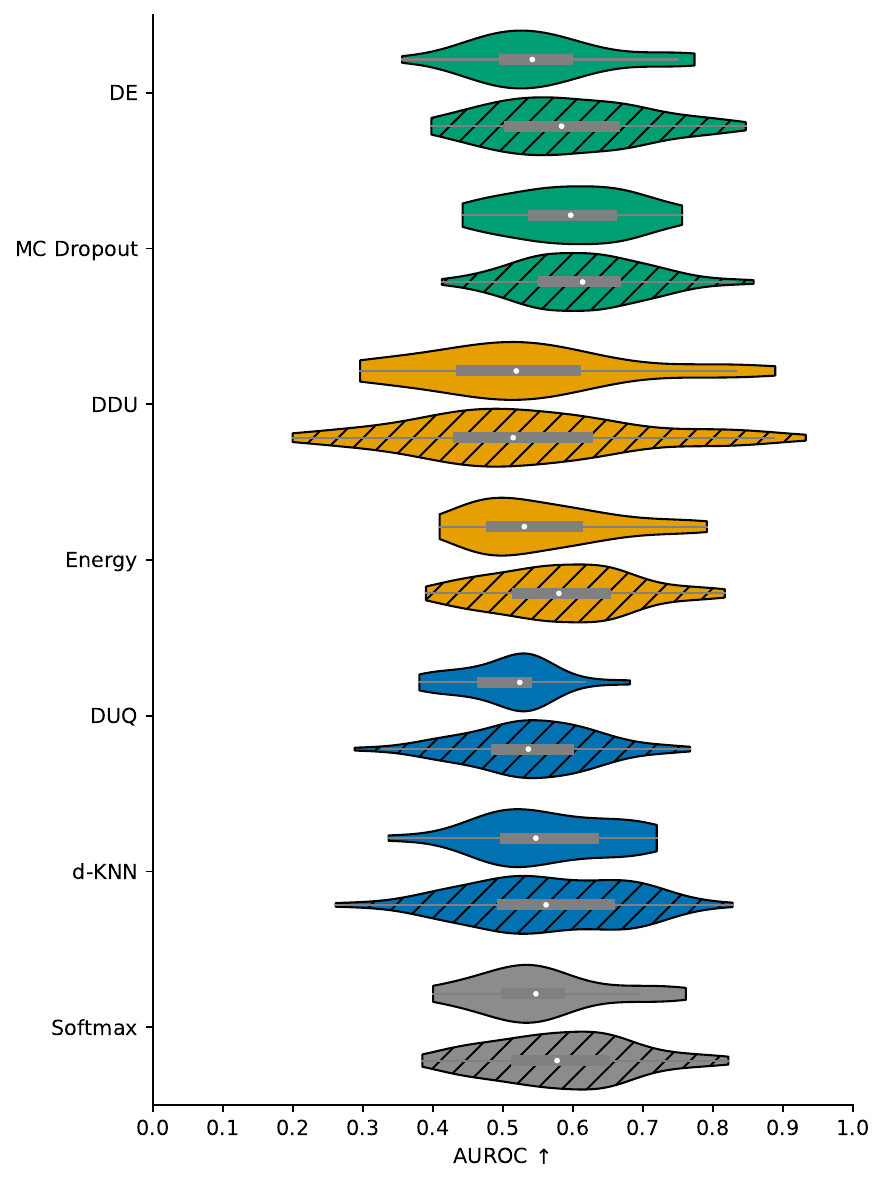}
        \caption{BNCI2014}
        \label{fig:cls_bnci} 
    \end{subfigure}
    \hfill
        \begin{subfigure}[t]{0.32\textwidth}
        \includegraphics[width=\linewidth]{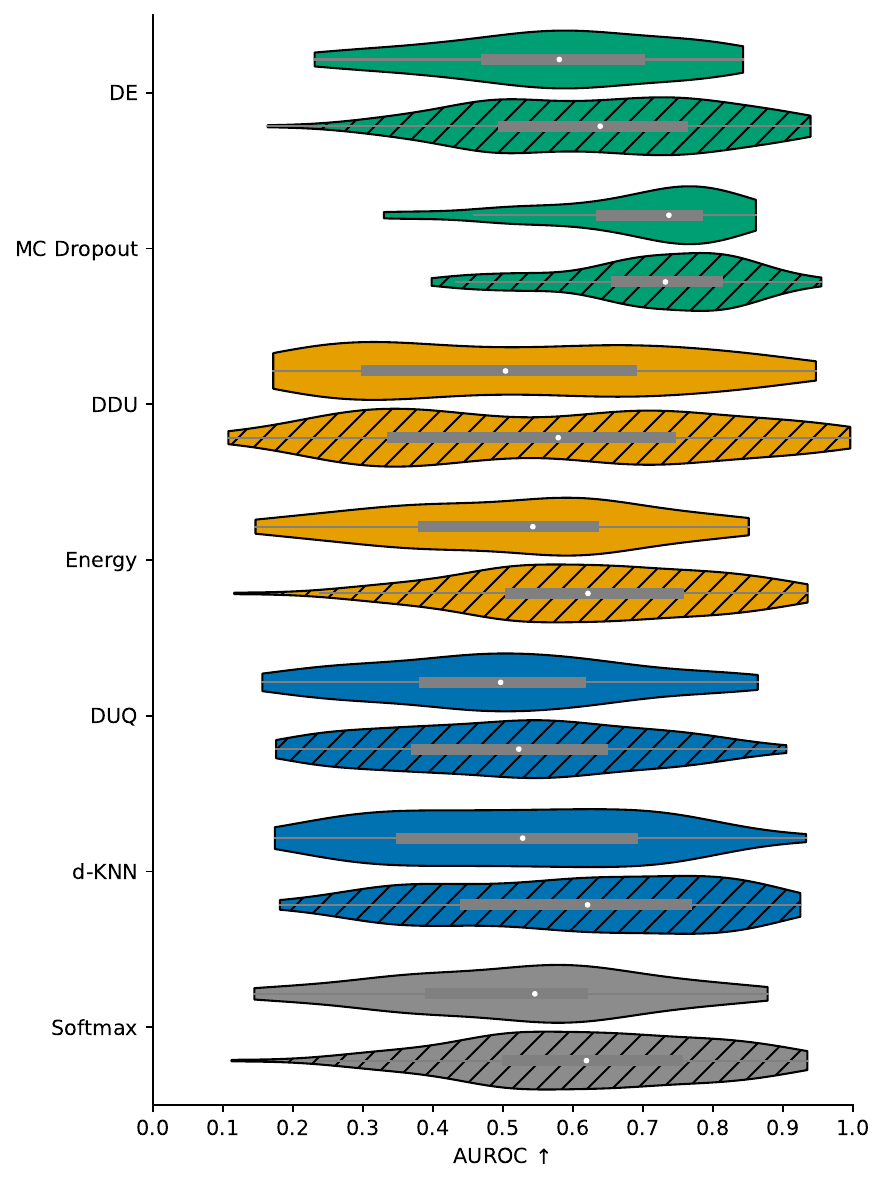}
        \caption{Schirrmeister2017}
        \label{fig:cls_schirrmeister} 
    \end{subfigure}
    \hfill
        \begin{subfigure}[t]{0.32\textwidth}
        \includegraphics[width=\linewidth]{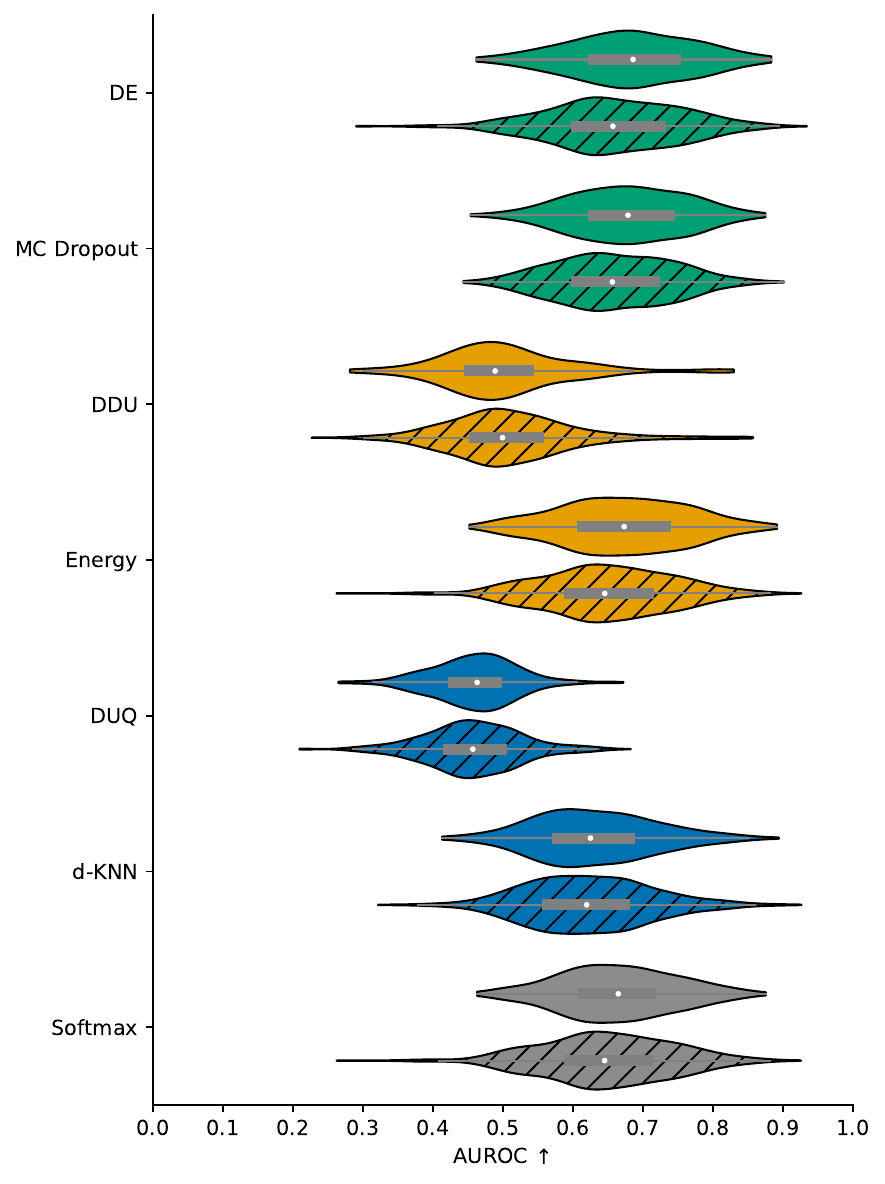}
        \caption{Stieger2021}
        \label{fig:cls_stieger} 
    \end{subfigure}
    
    \caption{The difference between OOD detection when a model is trained on three classes \tikzboxthree{} or two classes \tikzboxtwo{}. Colours indicate the following UQ categories: 
    \protect\legendbox{bayes}{Bayesian},
    \protect\legendbox{density}{Density},
    \protect\legendbox{distance}{Distance},
    \protect\legendbox{baseline}{Baseline}.}
    \label{fig:cls_diff}

\end{figure}

Figure \ref{fig:cls_schirrmeister}, displaying the results on the Schirrmeister2017 dataset, indicates that OOD detection may be easier when the model is trained on fewer classes. However, as shown in Table \ref{tab:test_results}, this effect is statistically significant only for the energy score on the Schirrmeister2017 dataset.
This pattern is inconsistent with the results for the Stieger2021 dataset (Figure \ref{fig:cls_stieger}). In that case, a significant difference is only observed for DE, energy score, and MC dropout, which perform better when the model is trained on three ID classes.
For the BNCI2014 dataset (Figure \ref{fig:cls_bnci}), we do not observe statistically significant differences.

\begin{table}[ht]
\centering
\begin{tabular}{l cc cc cc}
\toprule
\textbf{Method} & \multicolumn{2}{c}{\textbf{BNCI2014}} & \multicolumn{2}{c}{\textbf{Schirrmeister2017}} & \multicolumn{2}{c}{\textbf{Stieger2021}} \\
                & \textit{p} & \textit{Effect Size} & \textit{p} & \textit{Effect Size} & \textit{p} & \textit{Effect Size} \\
\midrule
Deep Ensemble  & 0.070 & 1866  & 0.058 & 4769  &  $\mathbf{8.440 \cdot 10^{-4}}$ & 76575 \\
MC Dropout     & 0.466 & 1669  & 0.335 & 4419  & $\mathbf{1.319 \cdot 10^{-3}}$ & 77056 \\
DDU            & 0.941 & 1550  & 0.192 & 4547  & 0.158 & 94683 \\
Energy         & 0.102 & 1834  & $\mathbf{7.250 \cdot 10^{-4}}$ & 5327  & $\mathbf{1.100 \cdot 10^{-3}}$ & 76858 \\
DUQ            & 0.044 & 1903  & 0.565 & 4272  & 0.576 & 87169 \\
d-KNN            & 0.919 & 1555  & 5.546 \(\cdot 10^{-3}\)& 5099  & 0.208 & 84506 \\
Softmax        & 0.081 & 1854  & 3.562 \(\cdot 10^{-3}\) & 5152  & 0.035 & 81245 \\
\bottomrule
\end{tabular}
\caption{Comparison of p-values and effect sizes from a Mann–Whitney U test across the three datasets for different methods. Bonferroni corrected significance threshold is $\alpha=2.381\cdot 10^{-3}$.}
\label{tab:test_results}
\end{table}
\clearpage
\section{Statistical analysis}\label{ap:appendices}
\captionsetup[table]{position=bottom}

\subsection{OOD detection method comparison}
Table \ref{tab:lmm_AUROC} contains a summary of the Linear Mixed Effects Model fitted on the results of the LOCO experiment with softmax on the BNCI2014 dataset as the baseline. As seen in the Table, the UQ methods perform significantly better on the Stieger2021 dataset compared to the BNCI2014 dataset. There is no significant difference between BNCI2014 and Schirrmeister2017. Furthermore, DDU and DUQ perform significantly worse on the Stieger2021 dataset, while MC dropout performs significantly better on the Schirrmeister2017 dataset.

\begin{table}[htbp]
\centering
\adjustbox{width=\textwidth}{
\begin{tabular}{lrrrrrr}
\toprule
Term & Coef. & Std.Err. & $z$ & $P>|z|$ & [0.025 & 0.975] \\
\midrule
Intercept & \textbf{0.554} & 0.025 & 22.493 & $\mathbf{4.852\cdot10^{-112}}$ & 0.506 & 0.603 \\

Method[T.DDU] & -0.027 & 0.022 & -1.224 & 0.221 & -0.071 & 0.016 \\
Method[T.DUQ] & \textbf{-0.048} & 0.022 & -2.147 & \textbf{0.032} & -0.091 & -0.004 \\
Method[T.Deep Ensemble] & 0.000 & 0.022 & 0.010 & 0.992 & -0.043 & 0.044 \\
Method[T.Energy] & 0.003 & 0.022 & 0.131 & 0.896 & -0.041 & 0.047 \\
Method[T.d-KNN] & 0.010 & 0.022 & 0.434 & 0.665 & -0.034 & 0.053 \\
Method[T.MC Dropout] & 0.042 & 0.022 & 1.901 & 0.057 & -0.001 & 0.086 \\

Dataset[T.Schirrmeister2017] & 0.004 & 0.031 & 0.113 & 0.910 & -0.058 & 0.065 \\
Dataset[T.Stieger2021] & \textbf{0.109} & 0.026 & 4.170 & $\mathbf{3.042\cdot10^{-5}}$ & 0.058 & 0.161 \\

Method[T.DDU]:Dataset[T.Schirrmeister2017] & -0.025 & 0.028 & -0.891 & 0.373 & -0.081 & 0.030 \\
Method[T.DUQ]:Dataset[T.Schirrmeister2017] & \textbf{-0.064} & 0.028 & -2.248 & \textbf{0.025} & -0.119 & -0.008 \\
Method[T.Deep Ensemble]:Dataset[T.Schirrmeister2017] & 0.022 & 0.028 & 0.782 & 0.434 & -0.033 & 0.078 \\
Method[T.Energy]:Dataset[T.Schirrmeister2017] & -0.009 & 0.028 & -0.314 & 0.754 & -0.064 & 0.047 \\
Method[T.d-KNN]:Dataset[T.Schirrmeister2017] & \textbf{-0.056} & 0.028 & -1.965 & \textbf{0.049} & -0.111 & -0.000 \\
Method[T.MC Dropout]:Dataset[T.Schirrmeister2017] & \textbf{0.112} & 0.028 & 3.959 & $\mathbf{7.539\cdot10^{-5}}$ & 0.057 & 0.167 \\

Method[T.DDU]:Dataset[T.Stieger2021] & \textbf{-0.135} & 0.024 & -5.695 & $\mathbf{1.236\cdot10^{-8}}$ & -0.181 & -0.088 \\
Method[T.DUQ]:Dataset[T.Stieger2021] & \textbf{-0.157} & 0.024 & -6.617 & $\mathbf{3.675\cdot10^{-11}}$ & -0.203 & -0.110 \\
Method[T.Deep Ensemble]:Dataset[T.Stieger2021] & 0.019 & 0.024 & 0.805 & 0.421 & -0.027 & 0.065 \\
Method[T.Energy]:Dataset[T.Stieger2021] & 0.005 & 0.024 & 0.227 & 0.820 & -0.041 & 0.052 \\
Method[T.d-KNN]:Dataset[T.Stieger2021] & -0.042 & 0.024 & -1.794 & 0.073 & -0.089 & 0.004 \\
Method[T.MC Dropout]:Dataset[T.Stieger2021] & -0.027 & 0.024 & -1.134 & 0.257 & -0.073 & 0.020 \\

Group Var & 0.003 & 0.006 & & & & \\
\bottomrule
\end{tabular}}
\caption{Linear Mixed Effects Model Results (Dependent Variable: AUROC).}
\label{tab:lmm_AUROC}
\end{table}

\subsection{Pairwise comparison}

Table \ref{tab:pairwise} compares the UQ methods aggregated over the datasets. MC Dropout outperforms all other methods, while DDU and DUQ perform relatively similarly but worse than all other methods.

\begin{table}[ht]
\centering
\begin{tabular}{lrrrrr}
\toprule
Contrast & Estimate & Std.Err. & z & p & $p_{\text{adj}}$ \\
\midrule
DDU - DUQ & 0.041 & 0.010 & 4.142 & $3.438 \cdot 10^{-5}$ & $2.407 \cdot 10^{-4}$ \\
DDU - Deep Ensemble & -0.095 & 0.010 & -9.642 & $5.302 \cdot 10^{-22}$ & $7.953 \cdot 10^{-21}$ \\
DDU - Energy & -0.082 & 0.010 & -8.397 & $4.597 \cdot 10^{-17}$ & $5.976 \cdot 10^{-16}$ \\
DDU - d-KNN & -0.058 & 0.010 & -5.870 & $4.367 \cdot 10^{-9}$ & $3.930 \cdot 10^{-8}$ \\
DDU - MC Dropout & -0.151 & 0.010 & -15.428 & $1.058 \cdot 10^{-53}$ & $2.117 \cdot 10^{-52}$ \\
DDU - Softmax & -0.081 & 0.010 & -8.219 & $2.050 \cdot 10^{-16}$ & $2.460 \cdot 10^{-15}$ \\
DUQ - Deep Ensemble & -0.135 & 0.010 & -13.785 & $3.158 \cdot 10^{-43}$ & $6.000 \cdot 10^{-42}$ \\
DUQ - Energy & -0.123 & 0.010 & -12.539 & $4.574 \cdot 10^{-36}$ & $8.234 \cdot 10^{-35}$ \\
DUQ - d-KNN & -0.098 & 0.010 & -10.012 & $1.351 \cdot 10^{-23}$ & $2.161 \cdot 10^{-22}$ \\
DUQ - MC Dropout & -0.192 & 0.010 & -19.571 & $2.761 \cdot 10^{-85}$ & $5.799 \cdot 10^{-84}$ \\
DUQ - Softmax & -0.121 & 0.010 & -12.361 & $4.227 \cdot 10^{-35}$ & $7.186 \cdot 10^{-34}$ \\
Deep Ensemble - Energy & 0.012 & 0.010 & 1.246 & 0.2129 & 0.4641 \\
Deep Ensemble - d-KNN & 0.037 & 0.010 & 3.773 & $1.616 \cdot 10^{-4}$ & $9.694 \cdot 10^{-4}$ \\
Deep Ensemble - MC Dropout & -0.057 & 0.010 & -5.786 & $7.211 \cdot 10^{-9}$ & $5.769 \cdot 10^{-8}$ \\
Deep Ensemble - Softmax & 0.014 & 0.010 & 1.423 & 0.1547 & 0.4641 \\
Energy - d-KNN & 0.025 & 0.010 & 2.527 & 0.0115 & 0.0575 \\
Energy - MC Dropout & -0.069 & 0.010 & -7.032 & $2.042 \cdot 10^{-12}$ & $2.042 \cdot 10^{-11}$ \\
Energy - Softmax & 0.002 & 0.010 & 0.177 & 0.8591 & 0.8591 \\
MC Dropout - Softmax & 0.071 & 0.010 & 7.209 & $5.634 \cdot 10^{-13}$ & $6.197 \cdot 10^{-12}$ \\
MC Dropout - d-KNN & 0.094 & 0.010 & 9.559 & $1.195 \cdot 10^{-21}$ & $1.672 \cdot 10^{-20}$ \\
Softmax - d-KNN & 0.023 & 0.010 & 2.349 & 0.0188 & 0.0752 \\
\bottomrule
\end{tabular}
\caption{Pairwise contrast estimates with standard errors, z-scores, and Holm-adjusted p-values.}
\label{tab:pairwise}
\end{table}
\clearpage

\section{Formal description of LOCO OOD}\label{ap:Formalism}

To help distinguish LOCO-OOD detection from similar tasks, we provide a formal definition. We first define the Machine Learning problem by considering that there is a Random Variable $X \in \mathbb{R}^d$, from which we sample epochs of EEG data for each trial. We also consider that there exists an unknown function $f: X \to Y$, that represents the mapping of EEG data to a ground truth class label $Y \in \{1, \dots, k\}$. The task is then to learn a function approximation $\hat{f} \approx f$ based on the training/calibration epochs $\{(x_1, y_1), \dots, (x_n, y_n)\}$ sampled from the calibration distribution $P(X, Y)$. For inference, the model is exposed to a test sample $x' \sim X'$, with corresponding ground truth label $y' =f(x')$, which we aim to estimate through our model $\hat{f}(x')$. We assume that the distribution $P(X)$ from which the calibration samples are drawn is similar to the distribution of the test data $P(X')$, while the function $y = f(x)$ stays consistent.

We define a sample $(x', y')$ as `Out-of-Distribution' with respect to the calibration distribution if $P(X=x', Y=y') = 0$. In Leave-One-Class-Out OOD detection this is guaranteed because the unseen class $y' =k+1$ has $P(Y=y') =0$, and therefore also $P(X=x', Y=y') = 0$. This is consistent with the definition of `Semantic Shift' from \citet{yang_generalized_2024}. Alternative OOD detection tasks focus on `Covariate Shift', aiming to create OOD samples where $P(X=x')=0$ \cite{bomatter2026ood}. 
A model that performs well on LOCO OOD detection is able to indicate these samples as OOD, while still being able to perform the classification task.

OOD detection is different from OOD generalisation, which instead aims to create a generalising model where a covariate-shifted sample $P(X=x')=0$ still gets a correct inference $\hat{f}(x') = f(x') =y'$ \cite{zheng2025beatrix}. It differs from artifact detection because artifacts also occur in the calibration epochs $x \sim X$. The concept of LOCO-OOD can be extended to regression problems by having a target $Y \in \mathbb{R}$ with finite support. This can be created by excluding samples $(x, y)$ where $ y\in[a, b]$ from the training data. Semantic Shift OOD samples will then have $P(y=y') =0$ if $y \in [a, b]$. 
\clearpage
\section{Variations of DDU and DUQ}\label{ap:ablation}
As described in Section \ref{sec:methods}, there are minor differences between the formulation of DDU and DUQ used in the main results and the original formulations. We provide a comparison between the altered formulation and the original formulation.

\subsection{DDU}

\begin{figure}[h!]
    \centering
    \includegraphics[width=0.5\linewidth]{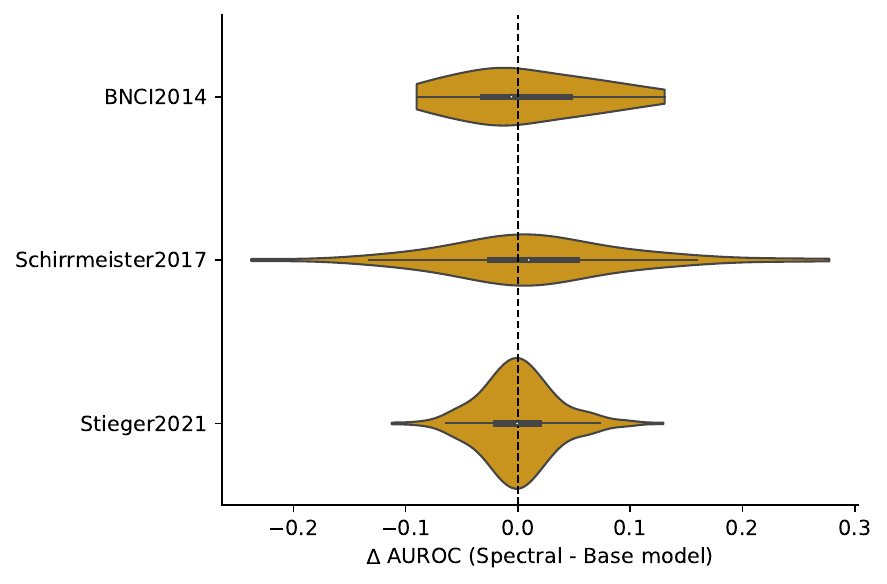}
    \caption{Difference in DDU's OOD detection performance when jointly applying spectral normalisation to all unconstrained convolutional layers and when only applying spectral normalisation to the penultimate convolutional layer (Base model).}
    \label{fig:ablation_ddu}
\end{figure}

Within our experiments, we tested DDU with a base model where spectral normalisation was applied to the penultimate layer only. The original DDU implementation is fitted on a fully constrained model \cite{mukhoti_deep_2023}. Therefore, we perform our LOCO experiment again where we apply spectral normalisation to all unconstrained convolutional layers in our base model. As seen in Figure \ref{fig:ablation_ddu}, we do not observe a consistent effect in either direction when comparing the difference in OOD detection AUROC. A Wilcoxon signed-rank test presented in Table \ref{tab:ablation_ddu} does not find any significant differences. When comparing the average on-task performance, we do not observe substantial differences either.

\begin{table}[h!]
    \centering
    \begin{tabular}{c cc cc}
    \toprule
            & \multicolumn{2}{c}{On-Task AUROC} & \multicolumn{2}{c}{Wilcoxon signed-rank} \\
    Dataset & $\mu$-Spectral & $\mu$-Base model & p value & effect size \\
    \midrule
        BNCI2014 &  0.830  & 0.837 & 0756 & 227  \\
        Schirrmeister2017   & 0.957 & 0.958 & 0.160 & 798  \\
        Stieger2021 & 0.912 & 0.912 &  0.508 & 14923 \\
    \bottomrule
    \end{tabular}
    \caption{Average on-task AUROC for both DDU implementations and results of the Wilcoxon signed-rank test on the OOD detection AUROC.}
    \label{tab:ablation_ddu}
\end{table}

\clearpage

\subsection{DUQ}

 Within our experiments, we train DUQ as a head on a frozen base model, whereas in the original implementation, DUQ is trained jointly with the base model \cite{amersfoort_uncertainty_2020}. Therefore, we perform our LOCO experiment again where we jointly train DUQ with an unfrozen base model.

\begin{figure}[h!]
    \centering
    \includegraphics[width=0.5\linewidth]{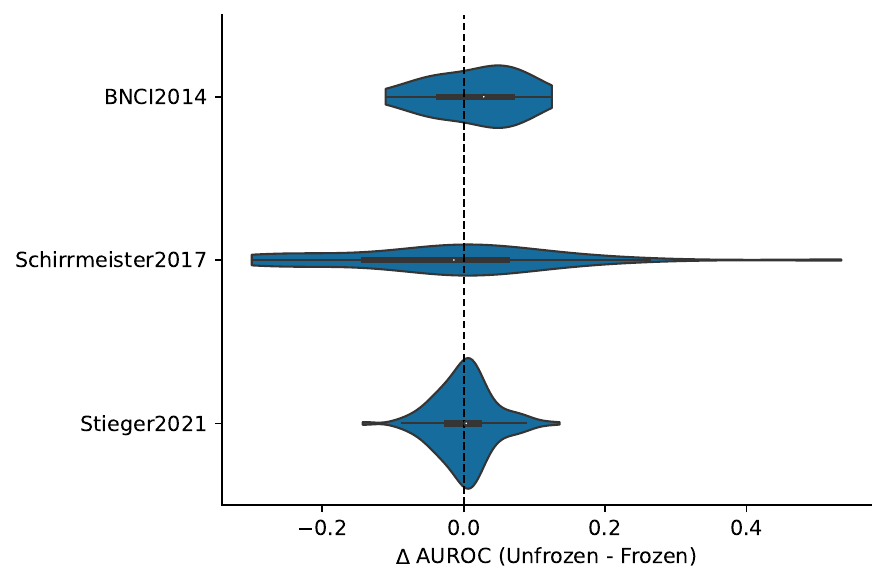}
    \caption{Difference in DUQ's OOD detection performance when jointly training DUQ with the feature extractor and when using a frozen, separately trained feature extractor.}
    \label{fig:ablation_duq}
\end{figure}

 Similar to the results for DDU, Figure \ref{fig:ablation_duq} shows no consistent effects on OOD detection, nor do we find any significant differences following a Wilcoxon signed-rank test presented in Table \ref{tab:ablation_duq}. 
 When comparing the average on-task performance, we do observe differences on the Schirrmeister2017 and the Stieger2021 datasets, where our frozen base model shows better on-task performance. 

\begin{table}[h!]
    \centering
    \begin{tabular}{c cc cc}
    \toprule
                & \multicolumn{2}{c}{On-Task AUROC} & \multicolumn{2}{c}{Wilcoxon signed-rank} \\
    Dataset & $\mu$-Unfrozen & $\mu$-Frozen & p & effect size \\
    \midrule
        BNCI2014 &  0.774  & 0.772 & 0.056 & 348  \\
        Schirrmeister2017   & 0.718 & 0.816 & 0.878 & 561\\
        Stieger2021 & 0.561 & 0.594 & 0.328 & 15438  \\
    \bottomrule
    \end{tabular}
    \caption{Average on-task AUROC for both DUQ implementations and results of the Wilcoxon signed-rank test on the OOD detect AUROC.}
    \label{tab:ablation_duq}
\end{table}
\clearpage
\section{Hyperparameters}\label{ap:Hyperparam}

Table \ref{tab:hyper_param_full} shows the hyperparameters used in the experiments for d-KNN and DUQ. Table \ref{tab:search_space} shows the search space in which the hyperparameters were found using the Optuna framework \cite{optuna_2019}. 
\begin{table}[h!]
\centering
\begin{tabular}{lcccc}
\toprule
Class & $k$ & Gamma ($\gamma$) & Centroid size & Penalty \\
\midrule

\addlinespace[0.8em]
\multicolumn{5}{c}{\textit{BNCI2014}} \\
\addlinespace[0.8em]

Left hand  & 17 & 0.989 & 32  & $1.33\times10^{-4}$ \\
Right hand & 29 & 0.998 & 64  & $3.78\times10^{-5}$ \\
Feet       & 13 & 0.990 & 128 & $2.83\times10^{-5}$ \\
Tongue     & 2  & 0.989 & 32  & $1.33\times10^{-4}$ \\

\addlinespace[0.8em]
\multicolumn{5}{c}{\textit{BNCI2014 two classes}} \\
\addlinespace[0.8em]

Left hand + Right hand & 19  & 0.913 & 64  & $1.19\times10^{-5}$ \\
Left hand + Feet       & 103 & 0.998 & 64  & $3.78\times10^{-5}$ \\
Left hand + Tongue     & 80  & 0.989 & 64  & $3.78\times10^{-5}$ \\

Right hand + Left hand & 59 & 0.995 & 128 & $1.20\times10^{-4}$ \\
Right hand + Feet      & 69 & 0.999 & 64  & $1.55\times10^{-3}$ \\
Right hand + Tongue    & 50 & 0.989 & 32  & $1.33\times10^{-4}$ \\

Feet + Left hand       & 105 & 0.998 & 64  & $3.78\times10^{-5}$ \\
Feet + Right hand      & 93  & 0.995 & 128 & $1.20\times10^{-4}$ \\
Feet + Tongue          & 33  & 0.989 & 64  & $1.19\times10^{-5}$ \\

Tongue + Left hand     & 73 & 0.998 & 64 & $3.78\times10^{-5}$ \\
Tongue + Right hand    & 73 & 0.998 & 64 & $3.78\times10^{-5}$ \\
Tongue + Feet          & 38 & 0.989 & 32 & $1.33\times10^{-4}$ \\

\addlinespace[0.8em]
\multicolumn{5}{c}{\textit{Schirrmeister2017}} \\
\addlinespace[0.8em]

Right hand & 23 & 0.998 & 64  & $3.78\times10^{-5}$ \\
Left hand  & 35 & 0.945 & 32  & $1.01\times10^{-4}$ \\
Rest       & 13 & 0.998 & 64  & $3.78\times10^{-5}$ \\
Feet       & 67 & 0.995 & 128 & $1.20\times10^{-4}$ \\

\addlinespace[0.8em]
\multicolumn{5}{c}{\textit{Schirrmeister2017 two classes}} \\
\addlinespace[0.8em]

Right hand + Left hand & 54 & 0.988 & 64  & $2.46\times10^{-4}$ \\
Right hand + Rest      & 62 & 0.992 & 64  & $7.59\times10^{-5}$ \\
Right hand + Feet      & 80 & 0.959 & 256 & $3.73\times10^{-2}$ \\

Left hand + Right hand & 60 & 0.995 & 128 & $1.20\times10^{-4}$ \\
Left hand + Rest       & 73 & 0.998 & 64  & $3.78\times10^{-5}$ \\
Left hand + Feet       & 52 & 0.998 & 128 & $3.51\times10^{-4}$ \\

Rest + Right hand      & 61 & 0.993 & 128 & $1.11\times10^{-3}$ \\
Rest + Left hand       & 79 & 0.990 & 128 & $2.83\times10^{-5}$ \\
Rest + Feet            & 75 & 0.995 & 128 & $1.20\times10^{-4}$ \\

Feet + Right hand      & 71 & 0.998 & 64  & $3.78\times10^{-5}$ \\
Feet + Left hand       & 59 & 0.989 & 32  & $1.33\times10^{-4}$ \\
Feet + Rest            & 75 & 0.993 & 32  & $5.23\times10^{-5}$ \\

\addlinespace[0.8em]
\multicolumn{5}{c}{\textit{Stieger2021}} \\
\addlinespace[0.8em]

Right hand & 1  & 0.996 & 32  & $2.23\times10^{-5}$ \\
Left hand  & 2  & 0.965 & 32  & $1.05\times10^{-3}$ \\
Both hands & 10 & 0.998 & 64  & $3.78\times10^{-5}$ \\
Rest       & 3  & 0.998 & 128 & $4.66\times10^{-3}$ \\

\addlinespace[0.8em]
\multicolumn{5}{c}{\textit{Stieger2021 two classes}} \\
\addlinespace[0.8em]

Right hand + Left hand  & 1  & 0.998 & 64  & $3.78\times10^{-5}$ \\
Right hand + Both hands & 20 & 0.998 & 64  & $3.78\times10^{-5}$ \\
Right hand + Rest       & 3  & 0.998 & 64  & $3.78\times10^{-5}$ \\

Left hand + Right hand  & 1  & 0.998 & 64  & $3.78\times10^{-5}$ \\
Left hand + Both hands  & 10 & 0.998 & 64  & $3.78\times10^{-5}$ \\
Left hand + Rest        & 4  & 0.998 & 64  & $3.78\times10^{-5}$ \\

Both hands + Right hand & 1  & 0.998 & 64 & $3.78\times10^{-5}$ \\
Both hands + Left hand  & 5  & 0.998 & 64 & $3.78\times10^{-5}$ \\
Both hands + Rest       & 14 & 0.998 & 64 & $3.78\times10^{-5}$ \\

Rest + Right hand       & 3  & 0.948 & 128 & $1.11\times10^{-4}$ \\
Rest + Left hand        & 35 & 0.999 & 64  & $1.55\times10^{-3}$ \\
Rest + Both hands       & 54 & 0.918 & 64  & $4.47\times10^{-5}$ \\

\bottomrule
\end{tabular}
\caption{Hyperparameters for d-KNN ($k$) and DUQ ($\gamma$, centroid size, penalty) across all datasets. Penalty refers to the magnitude of the gradient penalty during training.}
\label{tab:hyper_param_full}
\end{table}

\begin{table}[h!]
    \centering
    \begin{tabular}{lcc}
        \toprule
        Hyperparameter & Search space & Scale \\
        \midrule
        $\gamma$  & $[0.9, 0.999]$ & Log-uniform \\
        Centroid size                  & $\{32, 64, 128, 256\}$ & Categorical \\
        penalty          & $[10^{-5},\,5\times10^{-2}]$ & Log-uniform \\
        \bottomrule
    \end{tabular}
    \caption{Hyperparameter search space used during optimisation of the DUQ model.}
    \label{tab:search_space}
\end{table}
\end{appendices}

\end{document}